\documentclass{article}

\PassOptionsToPackage{table}{xcolor}
\usepackage[preprint]{neurips_2025}
\usepackage{parskip}
\usepackage{makecell}
\usepackage{sectsty} 
\usepackage{tcolorbox}
\usepackage[utf8]{inputenc} 
\usepackage[T1]{fontenc}    
\definecolor{mycolor}{RGB}{71,96,174}
\usepackage[colorlinks,linkcolor=mycolor,anchorcolor=mycolor,citecolor=mycolor,urlcolor=mycolor]{hyperref}   
\usepackage{url}            
\usepackage{booktabs}       
\usepackage{amsfonts}       
\usepackage{amsmath}        
\usepackage{float}          
\usepackage{nicefrac}       
\usepackage{microtype}      
\usepackage{graphicx}
\usepackage{enumitem}
\usepackage{multirow} 
\usepackage{wrapfig}
\usepackage{lipsum}
\usepackage{caption}
\usepackage{pifont}

\usepackage{fancyhdr}
\definecolor{lightblue}{rgb}{0.92, 0.96, 0.99}

\title{\raisebox{-0.5ex}{\includegraphics[height=3ex]{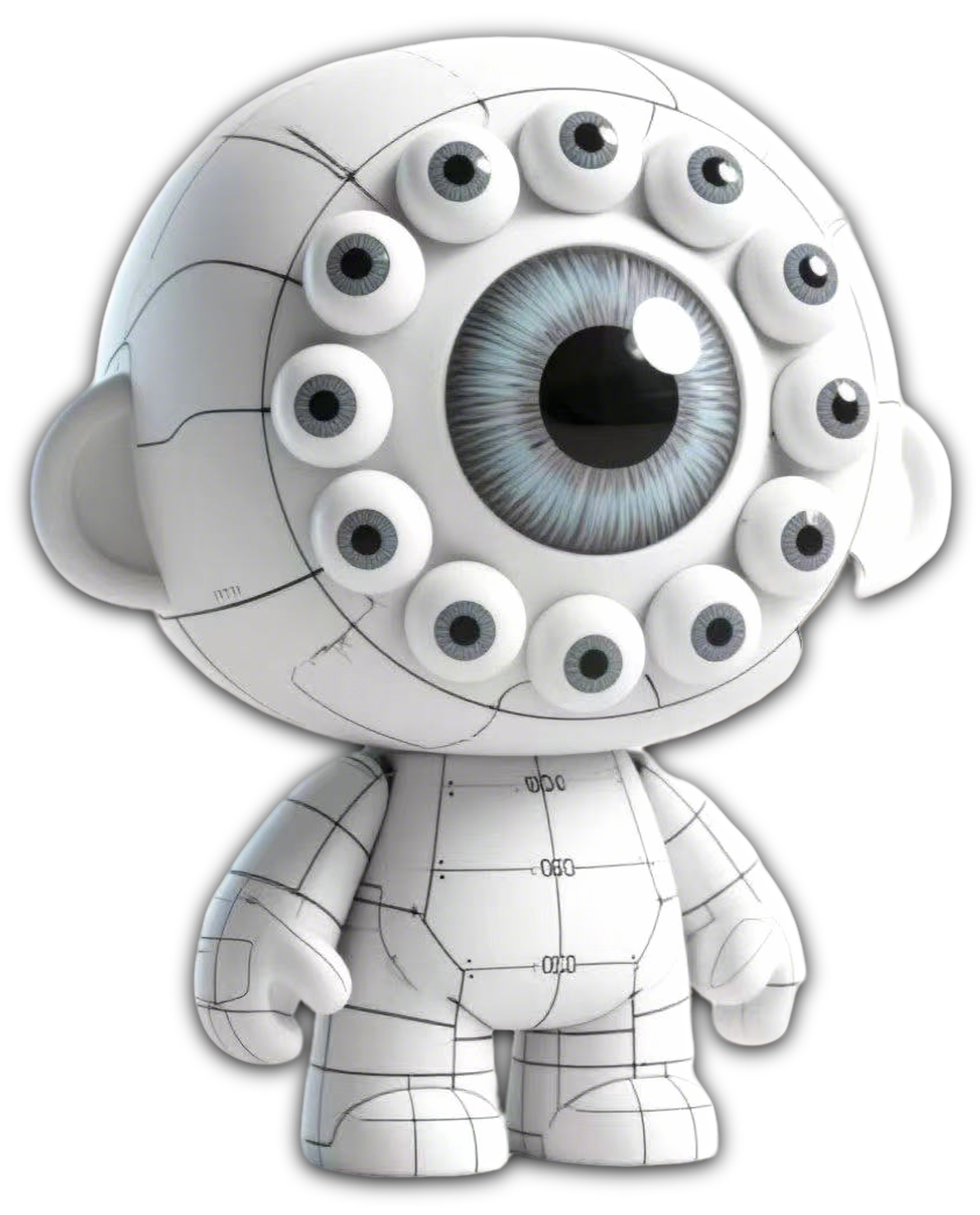}}~ArgusCogito: Chain-of-Thought for Cross-Modal Synergy and Omnidirectional Reasoning in Camouflaged Object Segmentation}

\author{
     Jianwen Tan$^{1,2}$$^*$, Huiyao Zhang$^{1,2}$$^*$, Rui Xiong$^{1,2}$, Han Zhou$^{1,2}$, Hongfei Wang$^2$, Ye Li$^2$$^{\dagger}$\\
    $^1$University of Chinese Academy of Sciences, \\ $^2$Technology and Engineering Center for Space Utilization, Chinese Academy of Sciences\\
     \quad$^*$Equal contributions  \quad$^{\dagger}$Corresponding author \\
     \url{https://zhang1huiyao.github.io/ArgusCogito/}\\
}

\begin{document}

\maketitle

\begingroup
\renewcommand{\thefootnote}{} 
\footnote{*Jianwen Tan and Huiyao Zhang contributed equally to this work. $^{\dagger}$Ye Li is the corresponding author(liye@csu.ac.cn).}
\addtocounter{footnote}{-1} 
\endgroup

\begin{abstract}
    Camouflaged Object Segmentation (COS) poses a significant challenge due to the intrinsic high similarity between targets and backgrounds, demanding models capable of profound holistic understanding beyond superficial cues. Prevailing methods, often limited by shallow feature representation, inadequate reasoning mechanisms and weak cross-modal integration, struggle to achieve this depth of cognition, resulting in prevalent issues like incomplete target separation and imprecise segmentation. Inspired by the perceptual strategy of the Hundred-eyed Giant—emphasizing holistic observation, omnidirectional focus, and intensive scrutiny—we introduce ArgusCogito, a novel zero-shot, chain-of-thought framework underpinned by cross-modal synergy and omnidirectional reasoning within Vision-Language Models (VLMs). ArgusCogito orchestrates three cognitively-inspired stages:
    (1) \textbf{\textit{Conjecture}}: Constructs a strong cognitive prior through global reasoning with cross-modal fusion (RGB, depth, semantic maps), enabling holistic scene understanding and enhanced target-background disambiguation.
    (2) \textbf{\textit{Focus}}: Performs omnidirectional, attention-driven scanning and focused reasoning, guided by semantic priors from Conjecture, enabling precise target localization and region-of-interest refinement.
    (3) \textbf{\textit{Sculpting}}: Progressively sculpts high-fidelity segmentation masks by integrating cross-modal information and iteratively generating dense positive/negative point prompts within focused regions, emulating Argus' intensive scrutiny.
    Extensive evaluations on four challenging COS benchmarks and three Medical Image Segmentation (MIS) benchmarks demonstrate that ArgusCogito achieves state-of-the-art (SOTA) performance, validate the framework's exceptional efficacy, superior generalization capability, and robustness.

\end{abstract}

\section{Introduction}

Camouflaged Object Segmentation (COS) confronts the critical challenge of delineating objects intrinsically concealed within their surroundings, where the core difficulty arises from extrinsic homogeneity between targets and backgrounds in color, texture, and structural patterns. Despite considerable progress, contemporary COS methodologies remain fundamentally constrained by two persistent limitations: (1) Inability to effectively identify all instances; (2) Failure to delineate the accurate boundaries of targets.

Current approaches exhibit significant shortcomings across dominant paradigms. Multimodal supervised methods (e.g., RGB-D fusion) leverage depth for geometric cues but lack semantic comprehension, reducing operations to naive feature concatenation rather than true understanding \cite{zongwei_2022}. Weakly-supervised techniques relying on sparse annotations (points/scribbles) reduce labeling costs at the expense of severe performance degradation in complex scenarios \cite{he2023weakly_b,he2023weakly_c}. Segment Anything Model (SAM) \cite{kirillov2023segment}-based solutions enable prompt-driven segmentation yet fail to encode task-specific semantics, resulting in inaccurate mask generation. While generic prompt strategies for Vision-Language Models (VLMs) eliminate manual annotation, they produce coarse localization that necessitates compensatory modular stacking (e.g., CLIP \cite{radford2021learning} + Grounding-DINO \cite{liu2024grounding}), introducing semantic fragmentation and information bottlenecks that impede end-to-end reasoning \cite{hu2024leveraging,hu2025int}.

Crucially, these approaches fail to synergistically integrate cross-modal information with the robust knowledge guidance and semantic reasoning capabilities inherent to VLMs. Methods that neglect the synergistic utilization of cross-modal information often struggle with accurate object localization and effective separation of objects from their backgrounds. Meanwhile, approaches that underutilize the semantic knowledge and reasoning capacities of VLMs tend to suffer from instance omission and imprecise boundary delineation, ultimately leading to poor generalization performance in complex scenarios.

To address these fundamental limitations, we introduce two key cognitive principles:

(1) \textbf{Cross-modal Synergy}: Unifying RGB, depth, and semantic priors to enhance target-background disambiguation via multimodal perception.

(2) \textbf{Omnidirectional Reasoning}: Emulating biological visual search through iterative, VLM-driven scrutiny for region refinement and semantic consistency.

\begin{figure}[t]
\centering
\includegraphics[width=1\columnwidth]{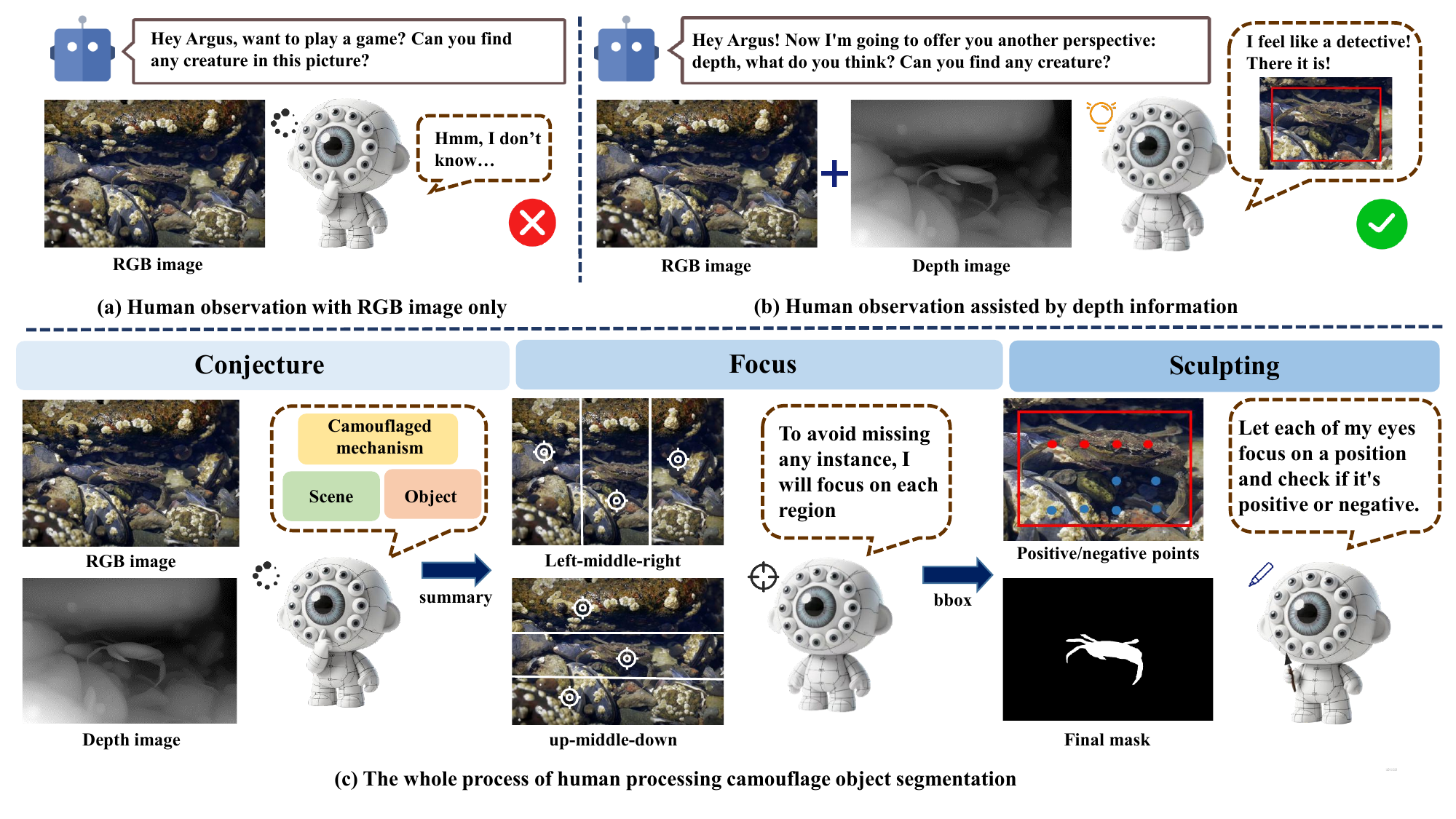} 
\caption{
Illustration of the human perceptual process in camouflaged object understanding, which inspires the design of our proposed ArgusCogito framework.  
}
\label{figure1}
\end{figure}

Inspired by the perceptual strategy of Argus Panoptes (the all-seeing giant in Greek mythology), we pioneer ArgusCogito, a zero-shot chain-of-thought framework that reformulates COS as a cohesive cognitive process. This framework orchestrates three biologically-motivated stages (as shown in Fig. 1c):

(1) \textbf{\textit{Conjecture}}: Constructs global cognitive priors via VLM-based fusion of RGB, depth, and semantic input, establishing holistic scene understanding for initial target-background disambiguation.

(2) \textbf{\textit{Focus}}: Performs VLM-driven omnidirectional attention scanning and focused reasoning, guided by semantic priors from Conjecture, achieving precise target localization and region-of-interest (ROI) refinement.

(3) \textbf{\textit{Sculpting}}: Progressively generates high-fidelity masks via iterative dense point prompting within ROIs, eliminating traditional CLIP dependencies by determining point polarity through VLM-based cross-modal verification (Argus-inspired intensive scrutiny).

As empirically validated in Figure~\ref{figure1}, ArgusCogito transcends conventional modular barriers by leveraging depth to reconstruct 3D perceptual priors , enabling lossless semantic transmission via unified VLM reasoning, and achieving unprecedented zero-shot generalization. Extensive evaluations across seven challenging benchmarks (three camouflaged animal datasets, one camouflaged plant dataset and three MIS datasets) confirm state-of-the-art performance, demonstrating exceptional efficacy, robustness, and cross-domain adaptability.

\section{Related Work}
\textbf{Multimodal Supervised Learning.}
Multimodal supervised learning methodologies enhance models’ perceptual capabilities in complex scenarios through the fusion of diverse modalities, including RGB images, depth maps, optical flow fields, and thermal images \cite{zongwei_2022,hongwei_2023,chenchen_2024,youjie_2024,junchi_2024}, thereby achieving improved detection accuracy and robustness. Depth information, in particular, plays a pivotal role by providing critical spatial separation cues between targets and backgrounds, facilitating the parsing of 3D scene structures by RGB modalities and enabling accurate discrimination of camouflaged objects \cite{qingwei_2023}. Certain approaches directly utilize RGB and depth data to enhance feature representation via cross-modal interactions \cite{chenghao_2025,hongbo_2024,xian_2023}. Alternative strategies focus on optimizing geometric features using depth maps, strengthening the perception of subtle structural differences through mechanisms such as boundary guidance \cite{hongbo_2023,youwei_2023}. However, these methods are constrained by their reliance on shallow feature correlations, failing to delve into the semantic logic inherent in prompts and thus struggling to form a comprehensive understanding of the target.

\textbf{Scribble/Point Prompt-Based Optimization.}
In complex visual scenarios, prompt mechanisms have emerged as a key approach to enhancing the robustness of segmentation and detection, owing to their capacity for fine-grained guidance of model attention—particularly in scenes characterized by blurred target boundaries or inconspicuous foregrounds. Existing prompt strategies can be broadly categorized: one class employs weakly supervised prompts (e.g., points, boundaries, scribbles) in conjunction with pseudo-label generation and structure-aware strategies to enable effective segmentation of camouflaged objects under weak supervision \cite{yu2024exploring,zhang2020weakly,he2023weakly_c,he2023weakly_b} . The second category explores automatic prompt generation via lightweight networks to minimize manual interaction, exemplified by AI-SAM \cite{yimu_2023}, which learns point weight distributions, and Self-Prompt-SAM \cite{bin_2025}, which extracts prompts from multi-scale masks. Notably, these methods lack integration with semantic information, leading to significant attenuation of prompt guidance in complex scenes, impairing their ability to sustain focus on target regions and ultimately undermining the robustness of segmentation and detection.

\textbf{Vision-Language Models.}
Contemporary methods frequently leverage the rich visual-language knowledge of multimodal large models (MLLMs/VLMs) for the recognition and segmentation of camouflaged targets \cite{junnan_2023,liu2023visual}. Specifically, Cheng et al.’s Multi-Level Knowledge Guidance (MLKG) approach decomposes complex tasks in a human-like manner to achieve strong generalization; Tang et al. developed a visual perception chain \cite{lv_2024} to enhance MLLM performance in camouflaged scenes through language prompt design and visual completion. Meanwhile, Hu et al. proposed cross-modal chain-of-thought prompting \cite{hu2024relax} to generate automatic visual and textual prompts; ProMaC \cite{hu2024leveraging}introduced an iterative prompt-mask framework that utilizes VLM hallucinations as prior knowledge alongside contrastive reasoning and INT \cite{hu2025int} employs progressive negative mining based on output changes induced by image occlusion . Despite performance gains in specific scenarios, these methods either rely on VLM fine-tuning or restrict VLMs to the localization stage, instead relying on modules such as CLIP \cite{radford2021learning} and Grounding-DINO \cite{liu2024grounding} for segmentation. This underutilization of VLMs’ powerful zero-shot knowledge guidance and semantic reasoning capabilities ultimately limits their performance in complex scenes.

\section{Methodology}
\begin{figure*}[tb]
\centering
\includegraphics[width=1\textwidth]{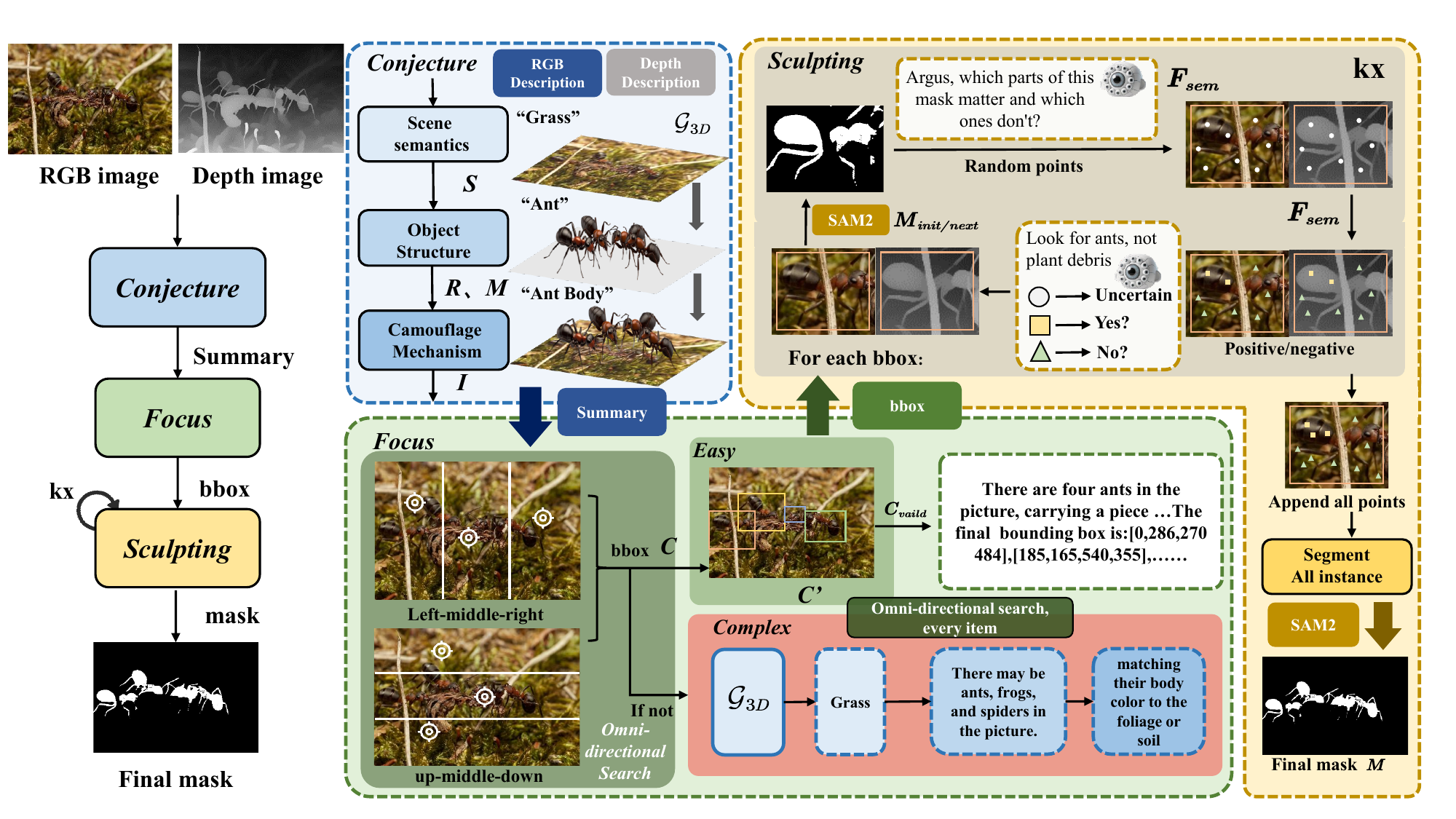}
\caption{
The architecture of \textit{ArgusCogito}, a zero-shot segmentation framework inspired by the perceptual strategy of the Hundred-eyed Giant. It comprises three cognition-driven stages: (1) \textit{Conjecture} builds a global prior via cross-modal fusion of RGB, depth, and semantics; (2) \textit{Focus} performs attention-guided, region-level reasoning for precise target localization; (3) \textit{Sculpting} refines masks iteratively using point-based prompts and semantic feedback. Together, these modules form a chain-of-thought reasoning pipeline within Vision-Language Models (VLMs).
}
\label{figure2}
\end{figure*}

We propose \textbf{ArgusCogito}, a zero-shot segmentation chain-of-thought applied to COS tasks. As illustrated in Figure~\ref{figure2}, high-quality masks are progressively generated through multimodal collaborative holistic perception, omnidirectional reasoning with focus, and iterative segmentation refinement. \textbf{ArgusCogito} is composed of three key modules—\textit{Conjecture}, \textit{Focus}, and \textit{Sculpting}. In the following sections, we detail the design and functionality of each module.

\subsection{Conjecture}  
To construct an overall cognition of scenes in camouflaged images, we propose \textit{Conjecture},  which leverages generic prompts to progressively guide Vision-Language Model (VLM) reasoning by synergizing multimodal information, thereby deconstructing the scenes, camouflaged objects, and their camouflage patterns. 

Given an RGB image ($I_{RGB}$) and its corresponding depth map ($I_{Depth}$), \textit{Conjecture} first constructs a global semantic representation $S$, which captures background layout, geometry, and environmental attributes. This is driven by a general scene-level prompt $P_{scene}$, allowing the VLM to interpret the spatial-semantic structure across both modalities:  
\begin{equation}\label{eqn:1}  
	S=\text{VLM}(P_{scene}, I_{RGB}, I_{Depth})  
\end{equation}  
where $S$ summarizes the scene's structure, incorporating camouflage-relevant factors like terrain,vegetation, and lighting.  

Using a general object-level prompt $P_{object}$, the VLM processes the same visual inputs alongside $S$, identifying candidate regions $R$ and generating semantic 3D structures $M$ corresponding to potential camouflaged targets:  
\begin{equation}\label{eqn:2}  
	R,M=\text{VLM}(P_{object}, I_{RGB}, I_{Depth}, S)  
\end{equation}  
where $R$ represents camouflaged objects likely detected in the scene, and $M$ refers to their inferred 3D shapes based on the 2D appearances from the current view. Through semantic alignment across perspectives, $M$ allows the model to predict the object’s morphology in other views, thus bridging 2D observations to 3D understanding.

To reason about how objects achieve camouflage in the given scene, \textit{Conjecture} activates a semantic reasoning prompt $P_{infer}$. This guides the VLM to infer plausible camouflage mechanisms based on the object structure $M$ and scene context $S$, generating a textual description $I$ that explains the observed camouflage behavior:  
\begin{equation}\label{eqn:3}  
	I=\text{VLM}(P_{infer}, M, S)  
\end{equation}  
where $I$ provides an interpretation linking object geometry with background semantics, explaining visual imperceptibility.  

The final output from \textit{Conjecture} is a unified feature set $\mathcal{G}_{3D}$, which serves as the foundational prior for the subsequent modules:  
\begin{equation}\label{eqn:4}  
	\mathcal{G}_{3D} = \{S, R, M, I\}  
\end{equation}

In complex or heavily occluded scenes, \textit{Conjecture} may fail to identify explicit camouflaged entities, instead providing coarse estimations of potential regions or abstract behavioral cues, which nonetheless provide essential guidance for the subsequent \textit{Focus} stage.

\subsection{Focus}
Although \textit{Conjecture} constructs an overall cognition of scenes and objects, its analysis of specific local details under complex circumstances remains insufficient, rendering further focused reasoning necessary for the accurate localization of camouflaged objects. To address this, we propose \textit{Focus}, a region-level semantic reasoning module designed to enable localized, cross-modal perception refinement.
\textit{Focus} performs omnidirectional, attention-driven scanning guided by $\mathcal{G}_{3D}$, and comprises four synergistic submodules: adaptive region decomposition, prompt-guided semantic focusing, hypothesis-driven reasoning, and candidate verification. 

\textit{Focus} integrates $\mathcal{G}_{3D}$, the RGB image ($I_{RGB}$), and the depth map ($I_{Depth}$) to form a comprehensive spatial-semantic representation. Utilizing pose-related semantics from $\mathcal{G}_{3D}$, geometric cues from $I_{RGB}$, and topological patterns from $I_{Depth}$, the module dynamically partitions the scene into $N$ subregions ($N=3$, optimizing granularity and semantic alignment). This adaptive decomposition, tailored to the target's inferred orientation (e.g., vertical: left--center--right; horizontal: top--center--bottom), is expressed as:
\begin{equation}\label{eqn:5}
	\{A_k\}_{k=1}^N = \text{VLM}(P_{div}, I_{RGB}, I_{Depth}, \mathcal{G}_{3D}),
\end{equation}
where $\{A_k\}$ represents the adaptive subregions, and $P_{div}$ is the region decomposition prompt.

Prompt-guided semantic focusing processes each subregion, fusing RGB--depth cues with $\mathcal{G}_{3D}$ to detect regions exhibiting semantic or modality-level inconsistencies indicative of camouflaged targets. These regions are delineated as candidate bounding boxes for subsequent refinement:
\begin{equation}\label{eqn:6}
	C = \text{VLM}(P_{foc}, \{A_k\}, I_{RGB}, I_{Depth}, \mathcal{G}_{3D}),
\end{equation}
where $P_{foc}$ is the focusing prompt, and $C$ denotes the initial candidate set.

For challenging camouflage scenarios where no candidates are detected, \textit{Focus} employs hypothesis-driven reasoning. The module activates scene-specific semantic concepts from $\mathcal{G}_{3D}$ to generate camouflage hypotheses ($\mathcal{H}$), which articulate plausible target patterns under strong camouflage. These hypotheses then guide a supplementary cross-modal search aimed at recovering previously missed targets. To improve directional granularity and maximize the semantic coverage of the search process, we increase the subregion partitioning to $N=6$ in this stage. This allows the model to scan the scene from finer orientations and gather richer hypothesis-grounded cues:
\begin{equation}\label{eqn:7}
	\mathcal{H} = \text{VLM}(P_{hyp}, \mathcal{G}_{3D}),
\end{equation}
\begin{equation}\label{eqn:8}
	C' = \text{VLM}(P_{scan}, \{A_k\}, I_{RGB}, I_{Depth}, \mathcal{H}),
\end{equation}
where $P_{hyp}$ and $P_{scan}$ are prompts for hypothesis generation and supplementary scanning, respectively, and $C'$ represents additional candidate regions.

Finally, \textit{Focus} verifies the biological plausibility of candidate regions by integrating multimodal inputs with $\mathcal{G}_{3D}$ under a verification prompt ($P_{ver}$). Candidates semantically aligned with the background are discarded, while validated bounding boxes ($C_{valid}$) are forwarded for mask refinement:
\begin{equation}\label{eqn:9}
	C_{valid} = \text{VLM}(P_{ver}, C^\dagger, I_{RGB}, I_{Depth}, \mathcal{G}_{3D}),
\end{equation}
where $C^\dagger \in \{C, C'\}$ dynamically selects the optimal candidate set based on detection efficacy. This integrated mechanism facilitates robust and precise localization, further reinforcing \textit{ArgusCogito}'s cross-modal, zero-shot segmentation performance.

\subsection{Sculpting}
Owing to the absence of semantic information infusion, zero-shot camouflaged object segmentation tasks are frequently plagued by problems like semantic inconsistency of segmented objects and ambiguity in boundary determination. Inspired by human iterative attention to uncertain visual boundaries, and incorporating multimodal visual information, we propose the \textit{Sculpting} —a language-guided iterative refinement mechanism for pixel-level mask enhancement via multi-round semantic feedback.

The \textit{Sculpting} progressively refines coarse initial masks into high-fidelity segmentations through multi-stage semantic correction. It integrates three mechanisms: semantic discrepancy assessment, point-guided structural adjustment, and iterative visual-semantic verification, which system guided by multimodal reasoning.

\textit{Sculpting} execution begins with candidate bounding boxes $C_{valid}$ from the \textit{Focus}. SAM2 \cite{ravi2024sam} generates an initial coarse mask $M_{\text{init}}$ for each region:
\begin{equation}\label{eqn:10}
	M_{\text{init}} = \text{SAM2}(I_{\text{RGB}}, I_{\text{Depth}}, C_{\text{valid}})
\end{equation}
While $M_{\text{init}}$ offers robust coverage, camouflage characteristics (e.g., texture blurring, discontinuous edges) often induce structural inaccuracies, leading to over- or under-segmentation.

To mitigate these deviations, \textit{Sculpting} employs the VLM to generate concise, instruction-style textual feedback. The VLM examines the mask’s structure and outputs short descriptions of its deficiencies—for example, noting “unclear edges,” “missing central coverage,” or “irregular shape”—which directly inform point classification.
\begin{equation}\label{eqn:11}
    F_{\text{sem}} = \text{VLM}(P_{\text{eval}}, M_{\text{init}}, I_{\text{RGB}}, I_{\text{Depth}}, \mathcal{G}_{3D})
\end{equation}

where $F_{\text{sem}}$ denotes the semantic feedback generated under evaluation prompt $P_{\text{eval}}$, providing a diagnostic of the mask’s semantic reliability and structural completeness. This feedback serves as a semantic prior in the refinement loop.

Based on $F_{\text{sem}}$, \textit{Sculpting}  initiates the point-prompted mask reconstruction phase. Within the current mask $M_{\text{cur}}$ bounding box, a pre-defined set of 10 regularly arranged points $P$ is generated. Then, under prompt $P_{\text{gen}}$, the VLM leverages its semantic understanding and knowledge from $\mathcal{G}_{3D}$ to categorize each point. Specifically, it identifies points corresponding to “biological core structures” as semantically representative positive points $P_{\text{pos}}$ to anchor the target's geometric center. Concurrently, points highly similar to the background are designated as negative points $P_{\text{neg}}$. Other points are discarded. This point classification and selection is formalized as:
\begin{equation}\label{eqn:12}
	\begin{aligned}
		P_{\text{pos}}, P_{\text{neg}} &= \text{VLM}(P_{\text{gen}}, P, M_{\text{cur}}, I_{\text{RGB}}, I_{\text{Depth}}, F_{\text{sem}}, \mathcal{G}_{3D})
	\end{aligned}
\end{equation}
Subsequently, SAM2 utilizes these prompts to produce a refined mask that better conforms to the target structure:
\begin{equation}\label{eqn:13}
	M_{\text{next}} = \text{SAM2}(I_{\text{RGB}}, I_{\text{Depth}}, P_{\text{pos}}, P_{\text{neg}})
\end{equation}
The newly generated mask $M_{\text{next}}$ then undergoes another round of semantic evaluation, iterating the refinement loop to progressively enhance boundary precision and semantic consistency. Based on our ablation study (see Table~\ref{table4}(b)), the number of iterations is empirically set to 3.

\section{Experiments}
\subsection{Experimental setup}

\textbf{Baselines.}
We evaluate ArgusCogito on three challenging domains: Camouflaged Object Segmentation (COS), Medical Image Segmentation (MIS), and Plant Camouflage Detection (PCD), all of which exhibit low semantic contrast and significant foreground–background ambiguity.
For COS, we compare with weakly supervised methods using point- and scribble-level annotations, and fully supervised RGB-D-based models. We also introduce a task-generic prompt setting, where only a textual description is provided without spatial priors.
Here,ArgusCogito incorporates Qwen2.5 with SAM2. Further experimentation in MIS and PCD tasks aims to validate our method’s superiority using task-generic prompts versus traditional techniques. We test combinations like GPT4V+SAM and LLaVA1.5+SAM to highlight the limitations of current VLM models in this context. Our ArgusCogito is also evaluated with leading state-of-the-art promptable segmentation methods to underline its effectiveness.  All results are averaged over three runs.

\textbf{Evaluation Metrics.}
Mean Absolute Error ($M$), adaptive F-value ($F_\beta$) \cite{margolin2014evaluate}, mean E-value ($E_\phi$) \cite{fan2021cognitive}, structural metric ($S_\alpha$) \cite{fan2017structure} and Weighted F-Measure ($F_{\beta}^w$) , where lower $M$ or higher $F_\beta$, $E_\phi$, $S_\alpha$ and $F_{\beta}^w$ indicate better performance.

\begin{table*}[tb]
	\centering
	\setlength\tabcolsep{1mm} 
	\fontsize{10pt}{11pt}\selectfont 
    \resizebox{\textwidth}{!}{
    	\begin{tabular}{c|c|cccc|cccc|cccc}
		\hline
		\multirow{2}{*}{Methods} & \multirow{2}{*}{Venue} &
		\multicolumn{4}{c|}{COD10K \cite{fan2021concealed} } &
		\multicolumn{4}{c|}{CHAMELEON \cite{skurowski2018animal} }  &
		\multicolumn{4}{c}{CAMO \cite{le2019anabranch} } \\
		\cline{3-6}
		\cline{7-10}
		\cline{11-14}
		& \multicolumn{1}{c|}{} & $M \downarrow$ & $F_{\beta} \uparrow$ & $E_{\phi} \uparrow$ & $S_{\alpha} \uparrow$ &
		$M \downarrow$ & $F_{\beta} \uparrow$ & $E_{\phi} \uparrow$ & $S_{\alpha} \uparrow$ &
		$M \downarrow$ & $F_{\beta} \uparrow$ & $E_{\phi} \uparrow$ & $S_{\alpha} \uparrow$ \\
        \hline
		\multicolumn{14}{c}{RGB-D-based COD Methods} \\
		\hline
        CDINet \cite{zhang2021cross} & MM21 & 0.044 & 0.610 & 0.821 & 0.778& 0.036 & 0.787 & 0.903 & 0.879 & 0.100 & 0.638 & 0.766 & 0.732  \\
		CMINet  \cite{zhang2021rgb} & CVPR21 & 0.039 & 0.768 & 0.868 & 0.811& 0.032 & \textbf{0.881} & 0.930 & 0.891 & 0.087 & 0.798 & 0.827 & 0.782  \\
		SPNet \cite{zhou2021specificity} & AAAI22 & 0.037 & 0.776 & 0.869 & 0.808& 0.033 & 0.872 & 0.930 & 0.888 & 0.083 & 0.807 & 0.831 & 0.783  \\
        DCMF \cite{wang2022learning} & TIP22 & 0.063 & 0.679 & 0.776 & 0.748& 0.059 & 0.807 & 0.853 & 0.830 & 0.115 & 0.737 & 0.757 & 0.728  \\
		DSAM \cite{yu2024exploring} & MM24 & 0.033 & 0.758 & 0.912 & 0.846 & 0.042 & 0.784 & 0.925 & 0.854 & 0.061 & 0.794 & 0.920 & 0.832 \\
        SAM-COD \cite{liu2025improving} & Arxiv25 & \textbf{0.022} & \textbf{0.827} & \textbf{0.948} & \textbf{0.887} & \textbf{0.029} & 0.850 & \textbf{0.961} & \textbf{0.893} & \textbf{0.044} & \textbf{0.849} & \textbf{0.952} & \textbf{0.875} \\
		\hline
		\multicolumn{14}{c}{Scribble Supervision Setting} \\
		\hline
		WSSA \cite{zhang2020weakly} & CVPR20 & 0.071 & 0.536 & 0.770 & 0.684 & 0.067 & 0.692 & 0.860 & 0.782 & 0.118 & 0.615 & 0.786 & 0.696 \\
		SCWS  \cite{yu2021structure} & AAAI21 & 0.055 & 0.602 & 0.805 & 0.710& 0.053 & 0.758 & 0.881 & 0.792 & 0.102 & 0.658 & 0.795 & 0.713  \\
		TEL \cite{zhang2020weakly} & CVPR22 & 0.057 & 0.633 & 0.826 & 0.724& 0.073 & 0.708 & 0.827 & 0.785 & 0.104 & 0.681 & 0.797 & 0.717  \\
		SCOD \cite{he2023weakly_c} & AAAI23 & 0.049 & 0.637 & 0.832 & 0.733& \textbf{0.046} & 0.791 & \textbf{0.897} & 0.818 & \textbf{0.092} & 0.709 & 0.815 & 0.735  \\
		SAM-S \cite{kirillov2023segment} & ICCV23 & 0.046 & 0.695 & 0.828 & 0.772& 0.076 & 0.729 & 0.820 & 0.650 & 0.105 & 0.682 & 0.774 & 0.731  \\
		WS-SAM \cite{he2023weakly_b}  & NeurlPS23 & \textbf{0.038} & \textbf{0.719} & \textbf{0.878} & \textbf{0.803} & \textbf{0.046} & \textbf{0.777} & \textbf{0.897} & \textbf{0.824} & \textbf{0.092} & \textbf{0.742} & \textbf{0.818} & \textbf{0.759} \\
		\hline
		\multicolumn{14}{c}{Point Supervision Setting} \\
		\hline
		WSSA \cite{zhang2020weakly} & CVPR20 & 0.148 & 0.607 & 0.652 & 0.649 & 0.087 & 0.509 & 0.733 & 0.642 & 0.105 & 0.660 & 0.712 & 0.711 \\
		SCWS \cite{yu2021structure} & AAAI21 & 0.142 & 0.624 & 0.672 & 0.687& 0.082 & 0.593 & 0.777 & 0.738& 0.097 &0.684 & 0.739 & 0.714   \\
		TEL \cite{zhang2020weakly} & CVPR22 & 0.133 & 0.662 & 0.674 & 0.645& 0.063 & 0.623 & 0.803 & 0.727& 0.094 & 0.712 & 0.751 & 0.746   \\
		SCOD \cite{he2023weakly_c} & AAAI23 & 0.137 & 0.629 & 0.688 & 0.663& 0.060 & 0.607 & 0.802 & 0.711 & 0.092 & 0.688 & 0.746 & 0.725  \\
		SAM \cite{kirillov2023segment} & ICCV23 & 0.160 & 0.597 & 0.639 & 0.643& 0.093 & 0.673 & 0.737 & 0.730& 0.207 & 0.595 & 0.647 & 0.635   \\
		SAM-P \cite{kirillov2023segment} & ICCV23 & 0.123 & 0.649 & 0.693 & 0.677& 0.069 & 0.694 & 0.796 & 0.765& 0.101 & 0.696 & 0.745 & 0.697   \\
		WS-SAM \cite{he2023weakly_b} & NeurlPS23 & \textbf{0.039} & \textbf{0.698} & \textbf{0.856} & \textbf{0.790}& \textbf{0.056} & \textbf{0.767} & \textbf{0.868} & \textbf{0.805} & \textbf{0.102} & \textbf{0.703} & \textbf{0.757} & \textbf{0.718}  \\
		\hline
		\multicolumn{14}{c}{Task-Generic Prompt Setting} \\
		\hline
		CLIP\_Surgery+SAM & Arxiv23 & 0.173 & 0.488 & 0.689 & 0.629& 0.147 & 0.606 & 0.741 & 0.689 & 0.189 & 0.520 & 0.692 & 0.612  \\
		GPT4V+SAM \cite{kirillov2023segment} & Arxiv23 & 0.187 & 0.448 & 0.672 & 0.601 & 0.180 & 0.557 & 0.710 & 0.637 & 0.206 & 0.466 & 0.666 & 0.573  \\
		LLaVA1.5+SAM \cite{liu2023visual} & NeurIPS23  & 0.170 & 0.530 & 0.728 & 0.662 & 0.168 & 0.561 & 0.718 & 0.666 & 0.314 & 0.401 & 0.585 & 0.501 \\
        \rowcolor{lightblue}
        Qwen2.5+SAM2 & Ours & 0.126 & 0.656 & 0.792 & 0.734 & 0.086 & 0.625 & 0.786 & 0.728 & 0.127 & 0.591 & 0.728 & 0.681 \\
		X-Decoder \cite{zou2023generalized} & CVPR23 & 0.171 & 0.556 & 0.705 & 0.652 & 0.124 & 0.654 & 0.748 & 0.716 & 0.104 & 0.628 & 0.745 & 0.709  \\
		SEEM \cite{zou2023segment} & NeurIPS23 & 0.143 & 0.001 & 0.280 & 0.425 & 0.094 & 0.011 & 0.307 & 0.454 & 0.192 & 0.023 & 0.315 & 0.404  \\
		GroundingSAM \cite{liu2024grounding} & ICCV23 & 0.085 & 0.670 & 0.813 & 0.764 & 0.122 & 0.662 & 0.776 & 0.744 & 0.157 & 0.656 & 0.753 & 0.707  \\
		GenSAM \cite{hu2024relax} & AAAI24 & 0.058 & 0.695 & 0.843 & 0.783 & 0.073 & 0.696 & 0.806 & 0.774 & 0.106 & 0.669 & 0.798 & 0.729  \\
		ProMaC \cite{hu2024leveraging} & NeurIPS24 & 0.042 & 0.716 & 0.876 & 0.805 & 0.044 & 0.790 & 0.899 & 0.833 & 0.090 & 0.725 & 0.846 & 0.767  \\
		INT \cite{hu2025int} & IJCAI25 & 0.037 & 0.722 & 0.883 & 0.808 & 0.039 & 0.801 & 0.906 & 0.842 & 0.086 & 0.734 & 0.853 & 0.772  \\
        \rowcolor{lightblue}
		ArgusCogito & Ours & \textbf{0.026} & \textbf{0.824} & \textbf{0.928} & \textbf{0.843}& \textbf{0.035} & \textbf{0.824} & \textbf{0.918} & \textbf{0.859} & \textbf{0.079} & \textbf{0.774} & \textbf{0.866} & \textbf{0.800}  \\
		\hline
	\end{tabular}
    }
	\caption{Results on Camouflaged Object Segmentation (COS) under different settings. Best are in bold}
    \label{table1}
\end{table*}

\textbf{PyTorch implementation details.}
We employ Qwen2.5-VL-7B-Instruct as the multimodal reasoning backbone to enable cross-modal understanding across all modules. For segmentation, the image encoder and mask decoder are derived from SAM 2.1 Hiera Large, initialized from official checkpoints. In \textit{Sculpting}, a binary point classifier adapted from SAM4MLLM verifies queried points for mask refinement.
Depth maps are generated from RGB inputs via Depth Anything. To simulate minimal prior knowledge under zero-shot conditions, we adopt a unified, task-generic prompt-based inference. Default prompts include “camouflaged animals” (COS), “camouflaged plants” (PCD), “polyp” and “skin lesion” (MIS).
All experiments are run on NVIDIA A100 GPUs. More architectural and training details are in the Appendix.

\subsection{Results and Analysis}
\textbf{Results on COS Tasks.}
The COS task focuses on precisely segmenting objects that are intentionally or naturally camouflaged within cluttered and visually deceptive. We evaluate our proposed model, \textit{ArgusCogito}, on three standard benchmarks: CHAMELEON (76 test images) \cite{skurowski2018animal}, CAMO (1,250 images: 1,000 training, 250 testing) \cite{le2019anabranch}, and COD10K (3,040 training, 2,026 testing images) \cite{fan2021concealed}. As presented in Table~\ref{table1}, we compare \textit{ArgusCogito} against diverse baselines: (1) fully supervised RGB-D models leveraging both RGB and depth modalities, (2) weakly supervised methods using point- or scribble-level annotations, and (3) vision-language segmentation frameworks operating under task-generic prompt settings without spatial supervision.

RGB-D models benefit from dense pixel-level annotations and geometric information, achieving strong performance under full supervision. Scribble-based methods generally outperform point-based approaches due to richer spatial cues. Despite operating without any form of supervision, \textit{ArgusCogito} consistently surpasses all weakly supervised methods and outperforms other state-of-the-art segmentation methods under the same prompt-driven, supervision-free paradigm. Notably, its performance rivals or exceeds several fully supervised RGB-D models, demonstrating robust zero-shot generalization.

This superior performance is attributed to our novel semantic reasoning framework, which integrates cross-modal signals and iteratively constructs internal spatial priors via a multimodal chain-of-thought process. Driven by task prompts, \textit{ArgusCogito} progressively refines predictions without relying on manual annotations or heuristic rules, enabling effective segmentation in visually challenging scenes.

\textbf{Results on MIS Tasks.}
To assess the zero-shot generalization of \textit{ArgusCogito}, we evaluate two representative medical image segmentation tasks—polyp segmentation (CVC-ColonDB \cite{tajbakhsh2015automated}, Kvasir \cite{jha2019kvasir}) and skin lesion segmentation (ISIC \cite{codella2019skin})—under a task-generic promptable setting without any fine-tuning. As shown in Table~\ref{table2}, \textit{ArgusCogito} achieves competitive zero-shot performance, exhibiting strong robustness to domain shift and input variation. Additional implementation details and visualizations are included in the Appendix.

\begin{table*}[h]
	\centering
	\setlength\tabcolsep{1mm} 
	\fontsize{10pt}{11pt}\selectfont 
    \resizebox{\textwidth}{!}{
        \begin{tabular}{c|c|cccc|cccc|cccc}
        \hline
        \multirow{3}{*}{Methods} & \multirow{3}{*}{Venue} & 
        \multicolumn{8}{c|}{Polyp Image Segmentation} & 
        \multicolumn{4}{c}{Skin Lesion Segmentation} \\
        \cline{3-14}
        & & 
        \multicolumn{4}{c|}{CVC-ColonDB \cite{tajbakhsh2015automated} } &
        \multicolumn{4}{c|}{Kvasir \cite{jha2019kvasir} } &
        \multicolumn{4}{c}{ISIC \cite{codella2019skin} } \\
        \cline{3-6} \cline{7-10} \cline{11-14}
        & & 
        $M \downarrow$ & $F_{\beta} \uparrow$ & $E_{\phi} \uparrow$ & $S_{\alpha} \uparrow$ &
        $M \downarrow$ & $F_{\beta} \uparrow$ & $E_{\phi} \uparrow$ & $S_{\alpha} \uparrow$ &
        $M \downarrow$ & $F_{\beta} \uparrow$ & $E_{\phi} \uparrow$ & $S_{\alpha} \uparrow$ \\
        \hline
		GroundingSAM \cite{liu2024grounding} & ICCV23 & 0.711 & 0.071 & 0.195 & 0.206 & 0.387 & 0.353 & 0.521 & 0.468 & 0.301 & 0.348 & 0.247 & 0.533 \\
		GenSAM \cite{hu2024relax} & AAAI24 & 0.244 & 0.059 & 0.494 & 0.379 & 0.172 & 0.210 & 0.619 & 0.487 & 0.171 & 0.699 & 0.744 & 0.678 \\
        ProMaC \cite{hu2024leveraging} & NeurIPS24 & 0.176 & 0.243 & 0.583 & 0.530 & 0.166 & 0.394 & 0.726 & 0.573 & 0.160 & 0.728 & 0.766 & 0.703 \\
        INT \cite{hu2025int}  & IJCAI25 & 0.172 & 0.250 & 0.589 & 0.537 & 0.161 & 0.401 & 0.732 & 0.574 & 0.152 & 0.733 & 0.771 & 0.708 \\
        \rowcolor{lightblue}
        ArgusCogito & Ours & \textbf{0.106} & \textbf{0.518} & \textbf{0.712} & \textbf{0.685} & \textbf{0.086} & \textbf{0.725} & \textbf{0.828} & \textbf{0.794} & \textbf{0.121} & \textbf{0.782} & \textbf{0.813} & \textbf{0.777} \\
		\hline
	\end{tabular}
    }
	\caption{Results for Medical Image Segmentation (MIS) under task-generic prompt setting}
    \label{table2}
\end{table*}

\textbf{Results on PCD Tasks.}
We further evaluate \textit{ArgusCogito} on the PlantCAMO \cite{yang2024plantcamo} dataset under a task-generic prompt setting. The Plant Camouflage Detection (PCD) task, formally introduced by Yang et al.~(2024), involves segmenting camouflaged plants in complex natural scenes. Comprehensive comparisons against both fully supervised and recent prompt-based approaches demonstrate that, without task-specific fine-tuning, \textit{ArgusCogito} consistently surpasses state-of-the-art task-generic baselines and performs on par with fully supervised models, as reported in Table~\ref{table3}. More experimental details and visualizations are provided in the Appendix.

\begin{table*}[h]
    \centering
    \setlength\tabcolsep{1mm}
    \fontsize{8.5pt}{9.5pt}\selectfont
    \begin{tabular}{c|c|cccc}
        \hline
        \multirow{2}{*}{Methods} & \multirow{2}{*}{Venue} &
        \multicolumn{4}{c}{PlantCAMO \cite{yang2024plantcamo}} \\
        \cline{3-6}
        & & $M \downarrow$ & $F_{\beta}^w \uparrow$ & $E_{\phi} \uparrow$ & $S_{\alpha} \uparrow$ \\
        \hline 
        DTINet \cite{liu2022boosting} & ICPR22 & 0.099 & 0.551 & 0.754 & 0.706 \\
        HitNet \cite{hu2023high} & AAAI23 & \textbf{0.081} & \textbf{0.618} & \textbf{0.784} & \textbf{0.740} \\
        VSCode \cite{luo2024vscode} & CVPR24 & 0.102 & 0.547 & 0.722 & 0.698 \\
        CamoDiffusion \cite{chen2024camodiffusion} & AAAI24 & 0.102 & 0.516 & 0.699 & 0.682 \\
        \hline
        GenSAM \cite{hu2024relax} & AAAI24 & 0.154 & 0.376 & 0.681 & 0.564 \\
        ProMaC \cite{hu2024leveraging} & NeurIPS24 & 0.137 & 0.437 & 0.726 & 0.600 \\
        \rowcolor{lightblue}
        ArgusCogito & Ours & \textbf{0.112} & \textbf{0.597} & \textbf{0.779} & \textbf{0.694} \\
        \hline
    \end{tabular}
    \caption{Quantitative results of SoTA on PlantCAMO.}
    \label{table3}
\end{table*}

\textbf{Module Analysis.}
We conduct ablation experiments on the CAMO dataset to investigate the impact of two core components in our \textit{Focus} and \textit{Sculpting} stages: the dynamic focus strategy and the number of segmentation iterations $k$.
To assess how spatial focus influences segmentation performance, we compare several configurations: \textit{single\_left} (left-center-right), \textit{single\_up} (top-center-bottom), \textit{double} (combining both directions), \textit{five} (covering left, right, top, bottom, and center), and our proposed \textit{auto} strategy that adaptively determines optimal focus regions.As shown in Table~\ref{table4}(a), \textit{single\_up} and \textit{single\_left} outperform the denser \textit{double} and \textit{five} setups, suggesting that overly redundant attention regions may introduce noise or overlap. In contrast, the \textit{auto} configuration achieves the best overall results, demonstrating the advantage of adaptive spatial reasoning for complex camouflage scenarios.We further examine the impact of iterative refinement in the \textit{Sculpting} stage by varying the number of iterations $k$ in the segmentation chain. Results in Table~\ref{table4}(b) indicate that increasing $k$ leads to consistent improvements, with optimal performance observed at $k=3$. This highlights the value of iterative reasoning and refinement for progressively enhancing segmentation precision in complex camouflage scenarios.

\begin{table*}[t]
    \centering
    \begin{minipage}{0.48\textwidth} 
        \centering
        \setlength\tabcolsep{1mm}
        \fontsize{8.5pt}{9.5pt}\selectfont

        \begin{tabular}{c|cccc}
        \hline
        \multirow{2}{*}{Methods} &
        \multicolumn{4}{c}{CAMO \cite{le2019anabranch}} \\
        \cline{2-5}
        & $M \downarrow$ & $F_{\beta} \uparrow$ & $E_{\phi} \uparrow$ & $S_{\alpha} \uparrow$ \\
        \hline
        single\_left & 0.081 & 0.758 & 0.862 & 0.797 \\
        single\_up & 0.079 & 0.761 & 0.863 & 0.797 \\
        double & 0.083 & 0.752 & 0.859 & 0.798 \\
        five & 0.085 & 0.746 & 0.862 & 0.795 \\
        \rowcolor{lightblue}
        auto & \textbf{0.079} & \textbf{0.774} & \textbf{0.866} & \textbf{0.800}\\
        \hline
        \end{tabular}

        \vspace{1mm}
        \textnormal{(a) Dynamic Focus Strategy.}
    \end{minipage}
    \hfill
    \begin{minipage}{0.48\textwidth} 
        \centering
        \setlength\tabcolsep{1mm}
        \fontsize{8.5pt}{9.5pt}\selectfont

        \begin{tabular}{c|cccc}
        \hline
        \multirow{2}{*}{k} &
        \multicolumn{4}{c}{CAMO \cite{le2019anabranch}} \\
        \cline{2-5}
        & $M \downarrow$ & $F_{\beta} \uparrow$ & $E_{\phi} \uparrow$ & $S_{\alpha} \uparrow$ \\
        \hline
        1 & 0.079 & 0.759 & 0.860 & 0.795 \\
        2 & 0.080 & 0.765 & 0.862 & 0.794 \\
        \rowcolor{lightblue}
        3 & \textbf{0.079} & \textbf{0.774} & \textbf{0.866} & \textbf{0.800} \\
        \hline
        \end{tabular}

        \vspace{1mm}
        \textnormal{(b) Number of Iterations $k$.}
    \end{minipage}
    \caption{Ablation study on CAMO dataset.}
    \label{table4}
\end{table*}

\begin{figure*}[tb]
\centering
\includegraphics[width=1\textwidth]{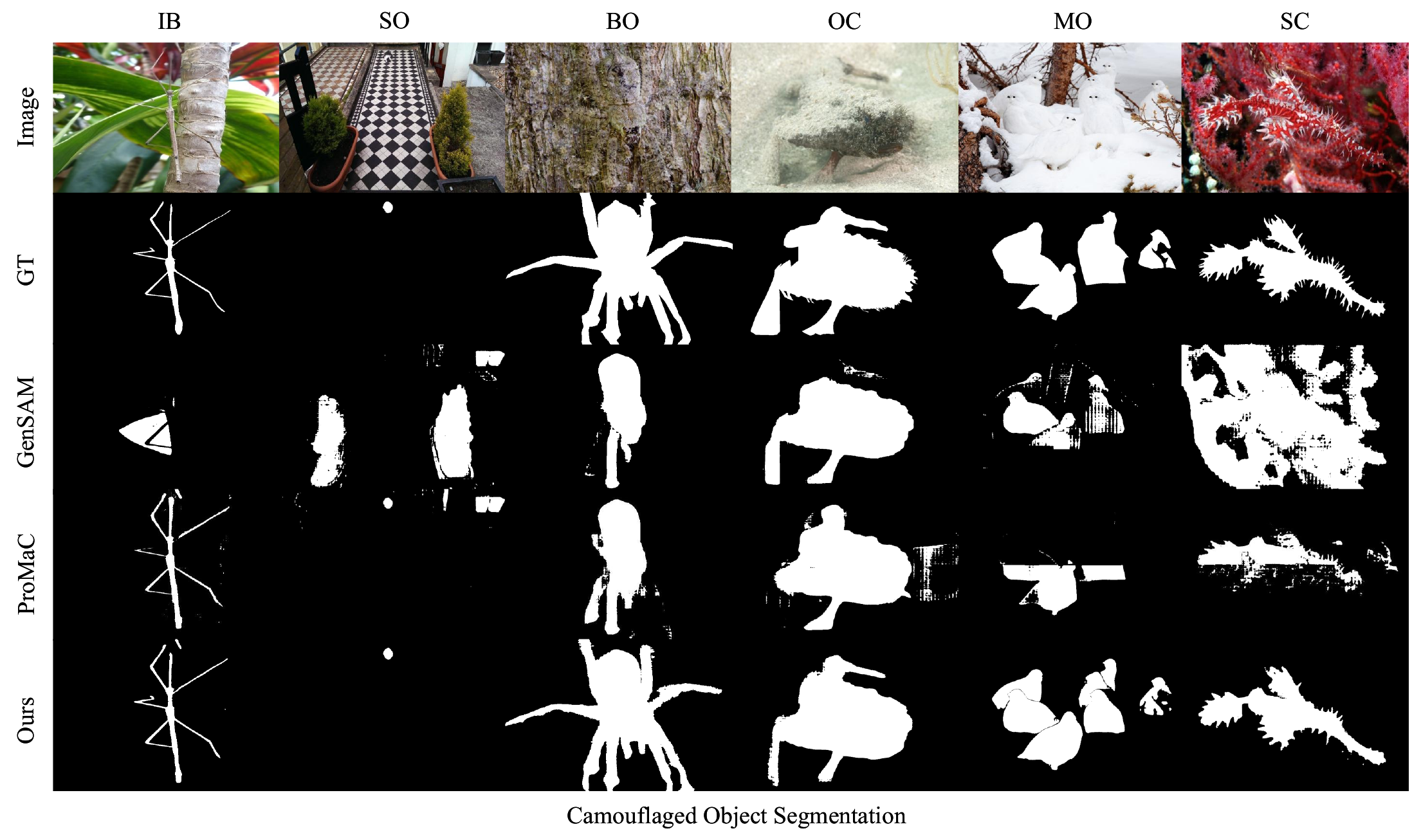} 
\caption{Qualitative visualization of camouflaged object segmentation (COS) across six challenging scenarios, including Indefinable Boundary (IB), Small Object (SO), Big Object (BO), Occlusion (OC), Multiple Objects (MO), and Shape Complexity (SC). The results compare the segmentation performance of ArgusCogito against GenSAM and ProMAC.}
\label{figure3_1}
\end{figure*}

\begin{figure*}[tb]
\centering
\includegraphics[width=1\textwidth]{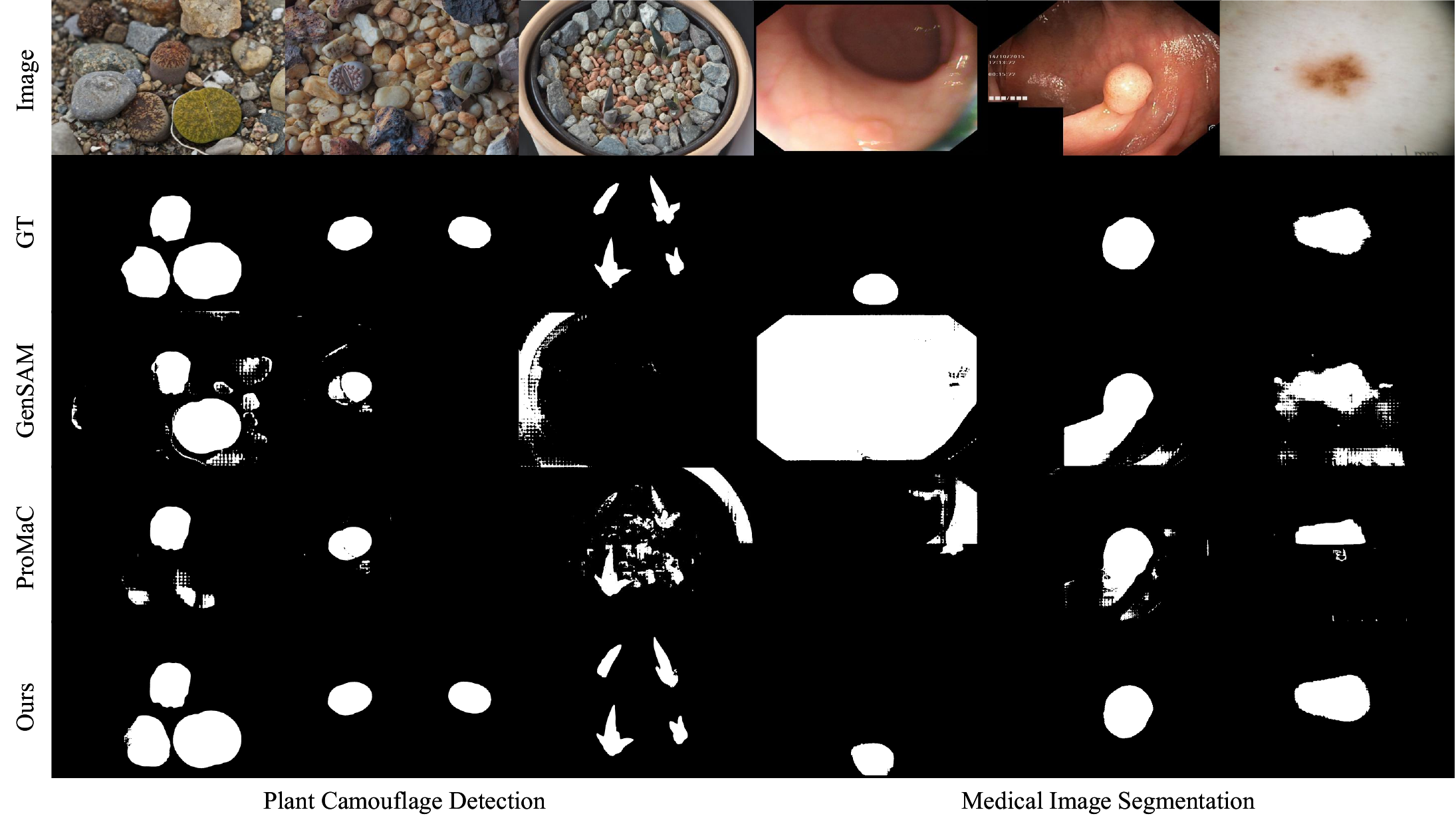} 
\caption{Visualization of medical image segmentation (MIS) and plant camouflage detection (PCD), comparing the performance of ArgusCogito against GenSAM and ProMAC.}
\label{figure3_2}
\end{figure*}

\textbf{Visual Analytics.}
Fig.~\ref{figure3_1} and Fig.~\ref{figure3_2} presents a qualitative comparison of our method with GenSAM and ProMAC across three domains: camouflaged object segmentation (COS), medical image segmentation (MIS), and plant camouflage detection (PCD). Our method consistently produces more precise masks with finer structures and clearer boundaries.
GenSAM derives visual prompts from generic text via cross-modal reasoning chains using BLIP2 and CLIP to generate consensus heatmaps. However, its outputs often suffer from spatial ambiguity and semantic drift, especially in cluttered scenes—resulting in false positives or misaligned masks.
ProMAC adopts an iterative hallucination-driven prompt-masking cycle based on multi-scale cropped patches. While enabling contextual reasoning, its reliance on independently processed views limits spatial completeness, frequently leading to under-segmentation when key regions lie outside cropped areas.
In contrast, our method performs holistic, full-image language-guided reasoning with iterative sculpting, ensuring robust semantic alignment and spatial coherence. Although INT, a prior state-of-the-art, is omitted due to unavailable code, our results show consistent superiority across COS, MIS, and PCD tasks, particularly under challenging conditions with ambiguous boundaries and complex textures.

\section{Conclusion}
In this work, we propose ArgusCogito, a novel zero-shot segmentation chain-of-thought that fully leverages the advantages of Vision-Language Models (VLMs). Through the synergy of cross-modal information, omnidirectional focused reasoning, and multi-round iterative mask optimization, it effectively achieves the conjecture, localization, and accurate segmentation of objects in scenes. Extensive experiments on Camouflaged Object Segmentation (COS), Medical Image Segmentation (MIS), and Plant Camouflage Detection (PCD) tasks demonstrate that ArgusCogito outperforms existing methods. It demonstrates the potential of zero-shot methods in addressing tasks such as camouflaged object segmentation, laying a foundation for future research on complex image segmentation across more domains.

\bibliographystyle{unsrt}
\bibliography{main.bbl}

\newpage
\appendix

\begin{center}
    {\LARGE \bf Supplementary Material}
\end{center}

\section*{Table of Contents}
\begin{itemize}

    \item Experimental Details \hfill \hyperref[sec:A]{Appendix A}
    \begin{itemize}
        \item Datasets Details \hfill \hyperref[sec:A1]{Appendix A.1}
        \item Implementation Details \hfill \hyperref[sec:A2]{Appendix A.2}
    \end{itemize}
    \item Additional Quantitative Evaluation \hfill \hyperref[sec:B]{Appendix B}
    \item Additional Ablation Studies \hfill \hyperref[sec:C]{Appendix C}
    \begin{itemize}
        \item Effect of Multimodal Backbone \hfill \hyperref[sec:C1]{Appendix C.1}
        \item Contribution of Depth Information \hfill \hyperref[sec:C2]{Appendix C.2}
        \item Prompt Sensitivity Analysis \hfill \hyperref[sec:C3]{Appendix C.3}
        \item Ablation Experiment Visualization \hfill \hyperref[sec:C4]{Appendix C.4}
    \end{itemize}
    \item Additional Qualitative Results and Qualitative Analysis \hfill \hyperref[sec:D]{Appendix D}
    \item Discussion \hfill \hyperref[sec:E]{Appendix E}
\end{itemize}

\section*{A Experimental Details}
\label{sec:A}
\subsection*{A.1 Datasets Details}
\label{sec:A1}
\subsubsection*{$\bullet$\ Camouflaged Object Segmentation Datasets}
In our experiments on camouflaged object perception, we adopt three representative datasets for camouflaged object detection: COD10K, CHAMELEON, and CAMO.
\begin{enumerate}
    \item \textbf{COD10K}: The largest and most comprehensive dataset for camouflaged object detection, containing 10,000 images categorized into 10 superclasses and 78 subclasses, 5,066 images contain camouflaged objects (3,040 for training and 2,026 for testing), accompanied by 1,934 non-camouflaged images and 3,000 background-only images. In the paper, we used 2026 images from its test set. 
    
    \item \textbf{CHAMELEON}: A small-scale dataset with 76 camouflaged images. It features complex natural backgrounds and manually annotated masks, serving to test generalization in unseen, high-complexity scenes.
    
    \item \textbf{CAMO}: Specifically designed for camouflaged object segmentation, comprising 2500 images (1250 camouflaged images from the internet + 1250 from MS-COCO). In the paper, we used the 250 images from its test set. Unlike other datasets, besides natural scenes, the test set of CAMO also includes body-painting images. Additionally, its lower resolution makes it more challenging compared to other datasets.
\end{enumerate}

\subsubsection*{$\bullet$\ Medical Image Segmentation Datasets}
In our experiments, the CVC-ColonDB and Kvasir-SEG datasets follow the dataset partitioning protocol proposed by PraNet, which has been widely adopted in recent medical image segmentation research. For the ISIC dataset, we use its official settings.
\begin{enumerate}
    \item \textbf{CVC-ColonDB}: A classic benchmark for colonic polyp detection, consisting of 300 still frames extracted from 15 colonoscopy videos, each containing a distinct polyp.In this paper, we utilize 380 testing samples.
    
    \item \textbf{Kvasir-SEG}: A high-quality dataset focused on polyp segmentation, encompassing diverse polyp shapes, sizes, textures, and realistic imaging artifacts, which present challenges for segmentation tasks.In this paper, we utilized the 100 images from its test set.
    
    \item \textbf{ISIC 2018}: Part of an international benchmark for skin lesion analysis, featuring rigorous protocols, cross-institutional test partitions, and metrics accounting for inter-observer variability. It serves as a standard for dermoscopic segmentation.In this paper, we utilized the 1000 images from its test set.
\end{enumerate}

\subsubsection*{$\bullet$\ Plant Camouflage Detection Datasets}
\begin{enumerate}
    \item \textbf{PlantCamo}: PlantCamo is a novel benchmark designed for plant camouflage detection (PCD). It uses high - resolution, copyright - free images sourced from platforms like Flickr. It features a diverse range of plant species, and its annotations cover both categories and instances. In this paper, we utilized the 250 images from its test set.
\end{enumerate}

\subsection*{A.2 Implementation Details}
\label{sec:A2}
\paragraph{$\bullet$\ Model and Module Setup}
\begin{enumerate}
    \item \textbf{Multimodal Reasoning Backbone} \\
    Qwen2.5-VL-7B-Instruct acts as the multimodal reasoning backbone, facilitating cross-modal semantic understanding across all modules of ArgusCogito. This foundation enables the system to interpret and fuse visual-language information effectively.
    
    \item \textbf{Segmentation and Sculpting Module} \\
    SAM 2.1 Hieral Large is employed for initial segmentation tasks. During the Sculpting stage, SAM4MLLM is introduced to perform point-level semantic verification. This refinement step checks if sampled points lie within target regions, enhancing mask accuracy and semantic consistency.
\end{enumerate}

\paragraph{ $\bullet$\ Workflow Execution}
\begin{enumerate}
    \item \textbf{Depth Map Pre-generation} \\
    For each dataset, Depth Anything-v2 is pre-executed to generate depth maps. These maps enrich spatial perception by providing dense geometric context, complementing RGB information and improving reasoning in cluttered or texture-weak scenes.
    
    \item \textbf{Chain-of-Thought Execution Modes} \\
    The conjecture-focus-sculpting pipeline supports two execution strategies for flexibility:
    \begin{enumerate}
        \item \textbf{Batch Step-by-Step Processing}: Modules (conjecture, focus, sculpting) run in batches, processing all images through one module before moving to the next. 
        \item \textbf{Sequential Single-Image Processing}: The full workflow (conjecture $\rightarrow$ focus $\rightarrow$ sculpting) completes for one image before starting the next. Our appendix provides code for this mode, simplifying debugging and single-image testing.
    \end{enumerate}
\end{enumerate}

\paragraph{$\bullet$\ Hardware and Optimization}
\begin{enumerate}
    \item \textbf{GPU Configuration} \\
    All experiments utilize NVIDIA A100 40G GPUs . Eight GPUs are grouped into four pairs (2 GPUs per pair) running in parallel. This setup accelerates large-scale inference (e.g., full-dataset segmentation) and ablation studies.
    
    \item \textbf{Memory Management for High-Resolution Inputs} \\
    To prevent out-of-memory errors during inference, images with excessively long edges are resized dynamically. The longest side is restricted to 1500 pixels, balancing resolution fidelity and computational feasibility.
\end{enumerate}

\section*{B Additional Quantitative Evaluation}
\label{sec:B}

Owing to space constraints in the main text, the comparative experimental results of certain methods in medical image segmentation tasks have not been included. We hereby provide supplementary information, with the additional methods encompassing GPT4V+SAM, LLaVA1.5+SAM, X-Decoder and SEEM. As summarized in Table~\ref{table5}.

The results validate ArgusCogito's strong zero-shot generalization capability in medical image segmentation, highlighting its potential as a domain-agnostic, cognitively inspired framework for high-ambiguity visual tasks.

\par\vspace{1em}
\noindent
\begin{minipage}{\columnwidth}
    \centering
    \setlength\tabcolsep{1mm} 
    \fontsize{10pt}{12pt}\selectfont 
    \resizebox{\textwidth}{!}{
    \begin{tabular}{c|c|cccc|cccc|cccc}
        \hline
        \multirow{3}{*}{Methods} & \multirow{3}{*}{Venue} & 
        \multicolumn{8}{c|}{Polyp Image Segmentation} & 
        \multicolumn{4}{c}{Skin Lesion Segmentation} \\
        \cline{3-14}
        & & 
        \multicolumn{4}{c|}{CVC-ColonDB} &
        \multicolumn{4}{c|}{Kvasir} &
        \multicolumn{4}{c}{ISIC} \\
        \cline{3-6} \cline{7-10} \cline{11-14}
        & & 
        $M \downarrow$ & $F_{\beta} \uparrow$ & $E_{\phi} \uparrow$ & $S_{\alpha} \uparrow$ &
        $M \downarrow$ & $F_{\beta} \uparrow$ & $E_{\phi} \uparrow$ & $S_{\alpha} \uparrow$ &
        $M \downarrow$ & $F_{\beta} \uparrow$ & $E_{\phi} \uparrow$ & $S_{\alpha} \uparrow$ \\
        \hline
        GPT4V+SAM & Arxiv23 & 0.578 & 0.051 & 0.246 & 0.242 & 0.614 & 0.128 & 0.236 & 0.253 & 0.514 & 0.387 & 0.366 & 0.334 \\
        LLaVA1.5+SAM  & NeruIPS23 & 0.491 & 0.194 & 0.355 & 0.357 & 0.479 & 0.293 & 0.400 & 0.403 & 0.369 & 0.473 & 0.497 & 0.477 \\
        X-Decoder & CVPR23 & 0.462 & 0.095 & 0.327 & 0.331 & 0.449 & 0.202 & 0.371 & 0.384 & 0.338 & 0.315 & 0.127 & 0.407 \\
        SEEM  & NeruIPS23 & 0.570 & 0.085 & 0.280 & 0.284 & 0.520 & 0.215 & 0.339 & 0.367 & 0.362 & 0.250 & 0.002 & 0.280 \\
        GroundingSAM  & ICCV23 & 0.711 & 0.071 & 0.195 & 0.206 & 0.387 & 0.353 & 0.521 & 0.468 & 0.301 & 0.348 & 0.247 & 0.533 \\
        GenSAM  & AAAI24 & 0.244 & 0.059 & 0.494 & 0.379 & 0.172 & 0.210 & 0.619 & 0.487 & 0.171 & 0.699 & 0.744 & 0.678 \\
        ProMaC  & NeurIPS24 & 0.176 & 0.243 & 0.583 & 0.530 & 0.166 & 0.394 & 0.726 & 0.573 & 0.160 & 0.728 & 0.766 & 0.703 \\
        INT & IJCAI25 & 0.172 & 0.250 & 0.589 & 0.537 & 0.161 & 0.401 & 0.732 & 0.574 & 0.152 & 0.733 & 0.771 & 0.708 \\
        ArgusCogito & Ours & \textbf{0.106} & \textbf{0.518} & \textbf{0.712} & \textbf{0.685} & \textbf{0.086} & \textbf{0.725} & \textbf{0.828} & \textbf{0.794} & \textbf{0.121} & \textbf{0.782} & \textbf{0.813} & \textbf{0.777} \\
        \hline
    \end{tabular}
    }
    \captionof{table}{Results for Medical Image Segmentation (MIS)}
    \label{table5}
\end{minipage}

\section*{C Additional Ablation Studies}
\label{sec:C}

\subsection*{C.1 Effect of Multimodal Backbone}
\label{sec:C1}
To investigate whether the performance gain stems primarily from the underlying foundation models or our proposed cognitive reasoning strategy, we conduct ablation studies that isolate the contribution of the multimodal backbone.

It is important to note that the architecture of ArgusCogito is intentionally tailored to the strengths of Qwen2.5-VL-7B, rather than relying solely on its capabilities. Qwen2.5 employs a patch-wise visual encoding strategy, which aligns well with our progressive reasoning paradigm—particularly the Focus stage that demands fine-grained spatial attention and localized decision-making. Additionally, Qwen2.5 benefits from diverse pretraining data, enabling it to support the Conjecture stage by constructing coarse yet semantically rich 3D conceptualizations of the scene. This property is critical for transferring geometric priors (e.g., from depth cues) without explicit supervision.

However, these advantages alone are insufficient to achieve state-of-the-art performance. As illustrated in Table~\ref{table6}, directly applying Qwen2.5-VL in conjunction with SAM2.1 without our reasoning framework results in significantly lower segmentation quality, particularly under complex camouflage conditions. This validates that the proposed cognitive chain—Conjecture, Focus, and Sculpting—is essential to unlocking the potential of the backbone, rather than merely aggregating pre-existing capabilities.

Moreover, the cognitive chain is not trivially transferable to other vision-language models. For example, LLaVA-1.5, despite supporting multimodal input, demonstrates poor generalization in both depth understanding and point-wise semantic verification. Unlike Qwen2.5, it lacks the fine-grained visual grounding and scene-level compositionality necessary to support our reasoning stages. Consequently, we do not include such models in the ablation, as they are fundamentally incompatible with the reasoning structure required by ArgusCogito.

In summary, the backbone model is a necessary enabler but not the primary contributor to performance. Our reasoning framework is specifically designed to amplify the visual-semantic alignment and spatial awareness of Qwen2.5-VL, leading to a synergistic effect that exceeds the sum of its parts. This design philosophy underscores the importance of co-designing task strategies with the capabilities of modern foundation models rather than treating them as plug-and-play components.

\par\vspace{1em}
\noindent
\begin{minipage}{\columnwidth}
	\centering
	\setlength\tabcolsep{1mm} 
	\fontsize{10pt}{12pt}\selectfont 
    \resizebox{\textwidth}{!}{
    \begin{tabular}{c|c|cccc|cccc|cccc}
		\hline
		\multirow{2}{*}{Methods} & \multirow{2}{*}{Venue} &
		\multicolumn{4}{c|}{COD10K} &
		\multicolumn{4}{c|}{CHAMELEON}  &
		\multicolumn{4}{c}{CAMO} \\
		\cline{3-6}
		\cline{7-10}
		\cline{11-14}
		& \multicolumn{1}{c|}{} & $M \downarrow$ & $F_{\beta} \uparrow$ & $E_{\phi} \uparrow$ & $S_{\alpha} \uparrow$ &
		$M \downarrow$ & $F_{\beta} \uparrow$ & $E_{\phi} \uparrow$ & $S_{\alpha} \uparrow$ &
		$M \downarrow$ & $F_{\beta} \uparrow$ & $E_{\phi} \uparrow$ & $S_{\alpha} \uparrow$ \\
        \hline
        Qwen2.5+SAM2 & Ours & 0.126 & 0.656 & 0.792 & 0.734 & 0.086 & 0.625 & 0.786 & 0.728 & 0.127 & 0.591 & 0.728 & 0.681 \\
		ArgusCogito & Ours & \textbf{0.026} & \textbf{0.824} & \textbf{0.928} & \textbf{0.843}& \textbf{0.035} & \textbf{0.824} & \textbf{0.918} & \textbf{0.859} & \textbf{0.079} & \textbf{0.774} & \textbf{0.866} & \textbf{0.800}  \\
		\hline
	\end{tabular}
    }
	\captionof{table}{Results on Camouflaged Object Segmentation (COS) under different settings.}
    \label{table6}
\end{minipage}

\subsection*{C.2 Contribution of Depth Information}
\label{sec:C2}
To further disentangle the contributions of spatial modalities in our framework, we investigate the role of depth information through targeted ablation. Specifically, we evaluate three configurations: (1) using only RGB inputs without depth or cognitive reasoning modules, (2) incorporating depth maps but disabling the ArgusCogito reasoning chain, and (3) leveraging both depth cues and our full Conjecture–Focus–Sculpting chain. This analysis allows us to isolate how depth contributes under varying levels of semantic interpretation and spatial abstraction.

Our findings reveal that depth alone offers limited and unstable benefits, as shown in Table~\ref{table7}. When depth maps are fused with RGB inputs without reasoning, segmentation performance improves only marginally on average. Notably, performance varies significantly across samples—depth improves results in images where geometric cues are pronounced (e.g., strong depth gradients or isolated foreground structures), but degrades in others where depth signals are weak, noisy, or spatially ambiguous. This variability results in inconsistent behavior and limited gains at the aggregate level, suggesting that raw geometric signals, without semantic grounding, are insufficient to resolve the ambiguity in camouflaged scenes.

In contrast, when depth is integrated with our structured reasoning process, it becomes a powerful enabler of spatial discrimination and high-level scene understanding. In the Conjecture stage, depth supports global reasoning by enabling coarse 3D scene conceptualization, allowing the model to infer object saliency despite suppressed visual cues. This facilitates the formation of robust object priors and the suppression of contextually implausible regions. In the Focus stage, depth enhances foreground-background disentanglement by introducing orthogonal spatial evidence, particularly beneficial in cluttered or low-contrast environments. In the Sculpting stage, depth plays a crucial role in guiding the iterative generation of positive and negative point prompts. These prompts help identify and correct geometrically inconsistent or low-confidence regions, ensuring structural coherence, especially for large, occluded, or fragmented objects.

These results highlight a key insight: depth is not merely an auxiliary feature but a context-dependent geometric scaffold that requires cognitive reasoning to be effectively utilized. Its utility is tightly coupled with the semantics of the reasoning chain; without such integration, depth can become unstable or even detrimental in complex scenes. The consistent performance gains observed under our full pipeline underscore the value of co-designing spatial priors and cognitive strategies. It is this synergy—rather than depth alone—that drives the superior robustness and accuracy observed in our camouflaged object segmentation results.

\par\vspace{1em}
\noindent
\begin{minipage}{\columnwidth}
	\centering
	\setlength\tabcolsep{1.2mm} 
	\fontsize{8.5pt}{9.5pt}\selectfont 
    \begin{tabular}{c|ccc|cccc}
		\hline
		\multirow{2}{*}{Setting} & \multicolumn{3}{c|}{Modules Enabled} &
		\multicolumn{4}{c}{COD10K} \\
		\cline{2-4} \cline{5-8}
		& RGB & Depth & ArgusCogito & $M \downarrow$ & $F_{\beta} \uparrow$ & $E_{\phi} \uparrow$ & $S_{\alpha} \uparrow$ \\
		\hline
		(1) & \ding{51} & \ding{55} & \ding{55} & 0.126 & 0.656 & 0.792 & 0.734 \\
		(2) & \ding{51} & \ding{51} & \ding{55} & 0.056 & 0.628 & 0.717 & 0.742 \\
		(3) & \ding{51} & \ding{51} & \ding{51} & \textbf{0.026} & \textbf{0.824} & \textbf{0.928} & \textbf{0.843} \\
		\hline
	\end{tabular}
	\captionof{table}{Ablation study on the COD10K dataset.}
    \label{table7}
\end{minipage}

\subsection*{C.3 Prompt Sensitivity Analysis}
\label{sec:C3}

In prompt-based multimodal reasoning systems, the formulation of prompts plays a critical role in shaping the model's perception, interpretation, and decision-making processes. Rather than serving as simple input triggers—as seen in conventional instruction-tuning paradigms—our framework treats prompts as semantic anchors that activate a structured Conjecture–Focus–Sculpting reasoning chain. To rigorously assess the robustness and sensitivity of this mechanism to variations in prompt formulation, we conduct a targeted ablation study across several domains and input modalities.

Each domain in our system is assigned a default prompt designed to encode not only high-level task semantics, but also latent spatial and structural priors necessary for effective zero-shot segmentation. For example, in the camouflaged object segmentation (COS) task, prompts such as “a camouflaged animal hidden in the scene” enable the model to hypothesize the presence of organisms that exhibit background-matching coloration, texture mimicry, or organic shapes. In the plant camouflage detection (PCD) domain, we employ prompts like “a camouflaged plant blending into the background,” which emphasize low-contrast boundaries and fractal-like shape structures. For medical image segmentation (MIS), the prompts are adapted to anatomical contexts—such as “a polyp inside the colonoscopy image” or “a skin lesion that is difficult to identify”—where structural irregularity and color consistency are more prominent cues.

To explore sensitivity, we construct multiple variants of each prompt by systematically manipulating the semantic granularity, lexical abstraction, and instructional specificity. These include overly verbose descriptive definitions, concise direct prompts, domain-general prompts (e.g., “find hidden objects”), and even underspecified natural-language requests (e.g., “Help me find a creature”). Furthermore, to examine how localized spatial guidance interacts with the Focus stage, we introduce directional cues (e.g., “analyze the upper part of the image for camouflaged regions”), which are particularly useful in cluttered or low-saliency scenarios.

Our experimental findings reveal several important insights. First, prompts that are semantically aligned with the target domain and incorporate structural or biological priors tend to yield the most stable and accurate segmentation outcomes. For example, prompts that explicitly reference biological patterns, such as symmetry or limb visibility, significantly improve both F-measure and E-measure scores. Second, prompts that encode reasoning rules—such as ensuring full-body coverage, minimizing fragmentation, or prioritizing completeness—achieve robust performance across diverse scenarios, demonstrating that clear operational grounding enhances reasoning consistency. Third, while concise and direct prompts (e.g., “please find the camouflaged object”) can achieve competitive results when well-aligned with the task, overly simplified or vague prompts exhibit marked performance drops, often due to failure in initiating effective spatial abstraction or semantic disambiguation. This trend is particularly evident in low-contrast, occluded, or geometrically ambiguous scenes.

Interestingly, the best-performing prompt in our study was a simplified Chinese directive that maintained high task specificity without overloading the model with linguistic noise. In contrast, casual or conversational prompts—though syntactically valid—produced significantly lower segmentation quality, underscoring the importance of semantic precision over surface fluency.

These findings highlight a central insight: the effectiveness of a prompt is not merely a function of its phrasing, but of its alignment with the underlying reasoning architecture. The cognitive structure of ArgusCogito is designed to leverage spatial, semantic, and contextual cues in a multi-stage process, and prompts that resonate with this structure unlock its full potential. Conversely, prompts that fail to establish meaningful anchors within the visual or semantic space impair the model’s ability to form hypotheses and reason effectively.

In summary, while our system exhibits a degree of robustness to moderate prompt variation, optimal performance is achieved when prompts are informative, structurally grounded, and task-aligned. This underscores the necessity of co-designing prompts in tandem with reasoning strategies, especially for zero-shot tasks in visually ambiguous or cognitively complex domains such as camouflaged object segmentation.

\begin{figure}[t]
\centering
\includegraphics[width=1\textwidth]{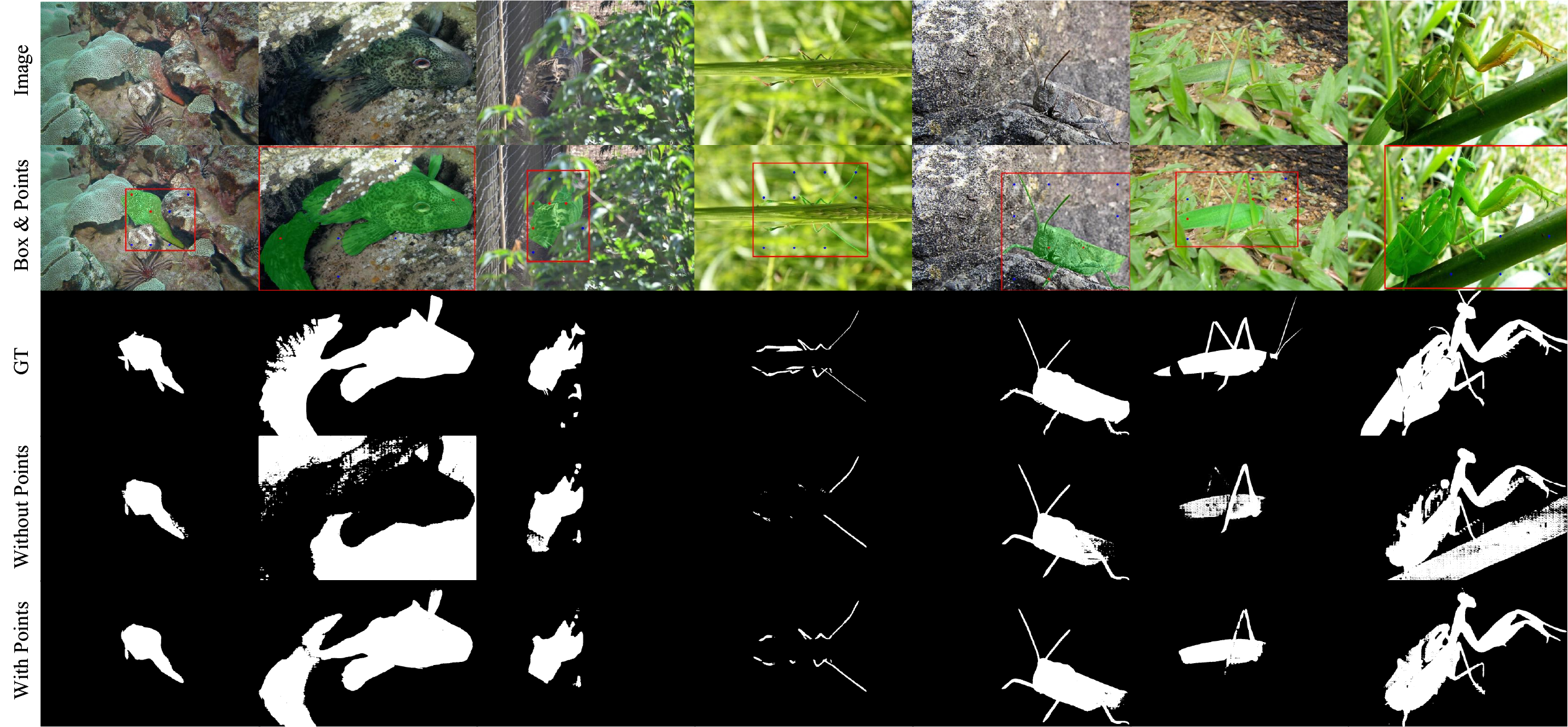}
\caption{Effect of removing point verification during the \textit{Sculpting} stage. Lack of semantic checking leads to spatial drift and boundary leakage.}
\label{figure4}
\end{figure}
\begin{figure}[t]
\centering
\includegraphics[width=0.9\textwidth]{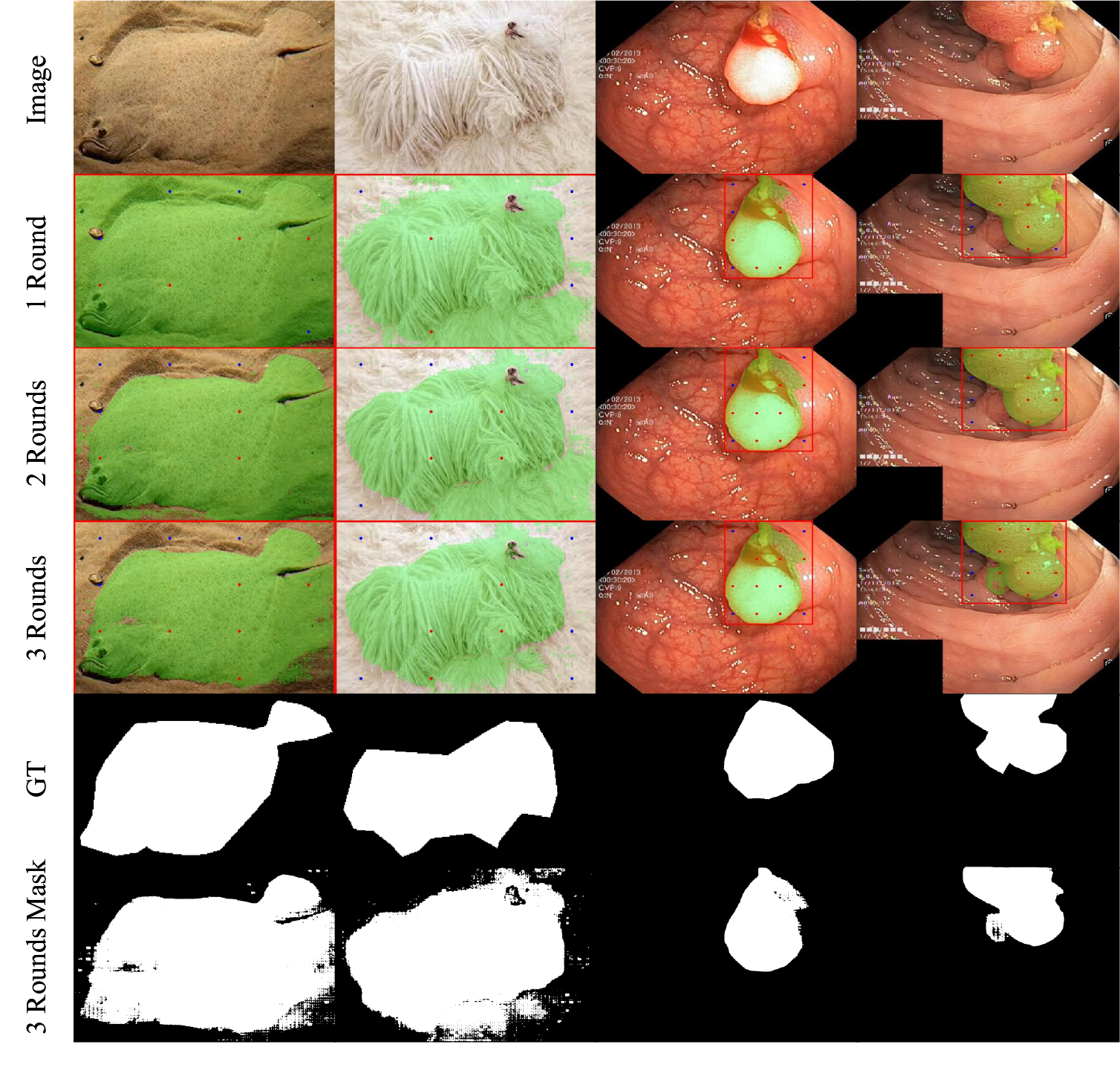}
\caption{Segmentation quality over three refinement rounds ($k=1,2,3$). Progressive reasoning yields finer masks and more accurate contours.}
\label{figure5}
\end{figure}

\subsection*{C.4 Ablation experiment visualization}
\label{sec:C4}

\textbf{Qualitative Analysis 1: The Impact of Point Verification  on Model Results}
Figure~\ref{figure4} visualizes segmentation results without point verification during the \textit{Sculpting} stage. The absence of binary point classification leads to spatial leakage and semantic drift, particularly near ambiguous boundaries. This highlights the importance of semantic point verification in reinforcing class consistency and boundary fidelity.

\textbf{Qualitative Analysis 2: The Impact of the Number of Iterations on Model Results}
Figure~\ref{figure5} shows the effect of iterative segmentation refinement. From the first to the third round, masks become progressively cleaner and more aligned with object contours, confirming that iterative sculpting facilitates robust refinement under weak supervision.

\section*{D Additional Qualitative Results and Qualitative Analysis}
\label{sec:D}

\begin{figure}[t]
\centering
\includegraphics[width=1\textwidth]{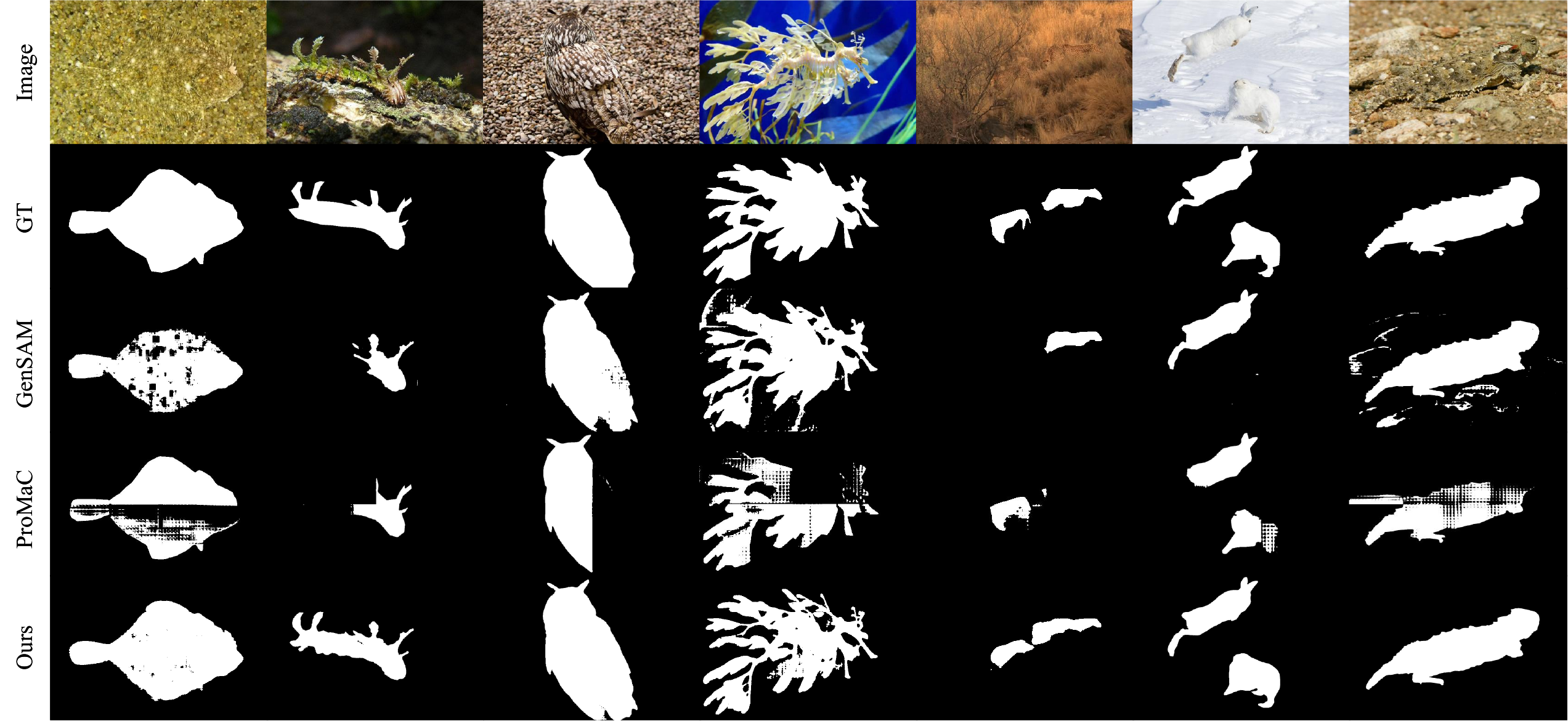}
\caption{Qualitative results on the CAMO dataset.}
\label{figure6}
\end{figure}
\begin{figure}[t]
\centering
\includegraphics[width=1\textwidth]{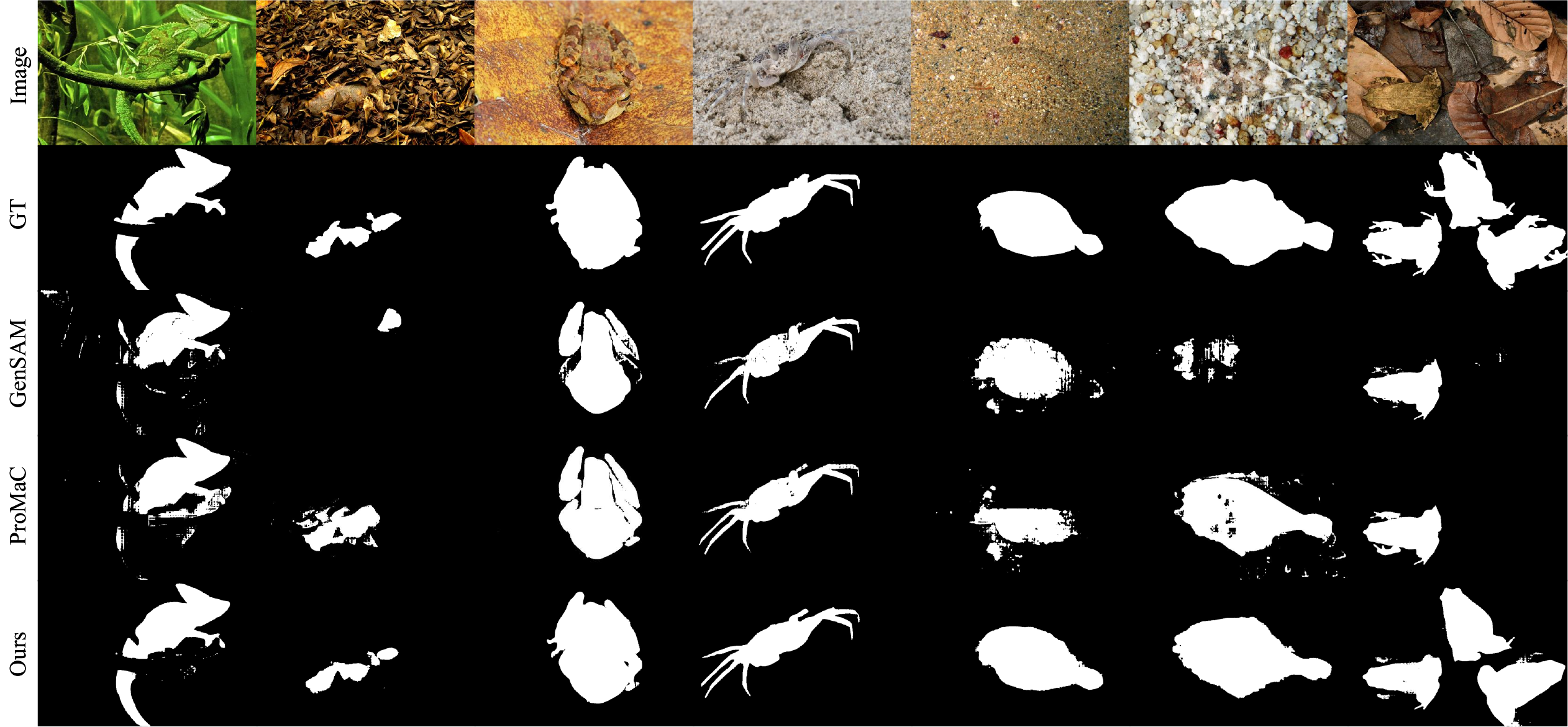}
\caption{Qualitative results on the CHAMELEON dataset.}
\label{figure7}
\end{figure}

To complement the main paper’s visual analysis and provide a more comprehensive understanding of \textit{ArgusCogito}'s segmentation behavior, we present additional qualitative results across five representative datasets: CAMO, CHAMELEON, COD10K, PlantCAMO and MIS task datasets. These datasets collectively span diverse challenges such as texture-background entanglement, semantic ambiguity, low inter-region contrast, and domain-specific occlusions.

\begin{figure}[tb]
\centering
\includegraphics[width=1\textwidth]{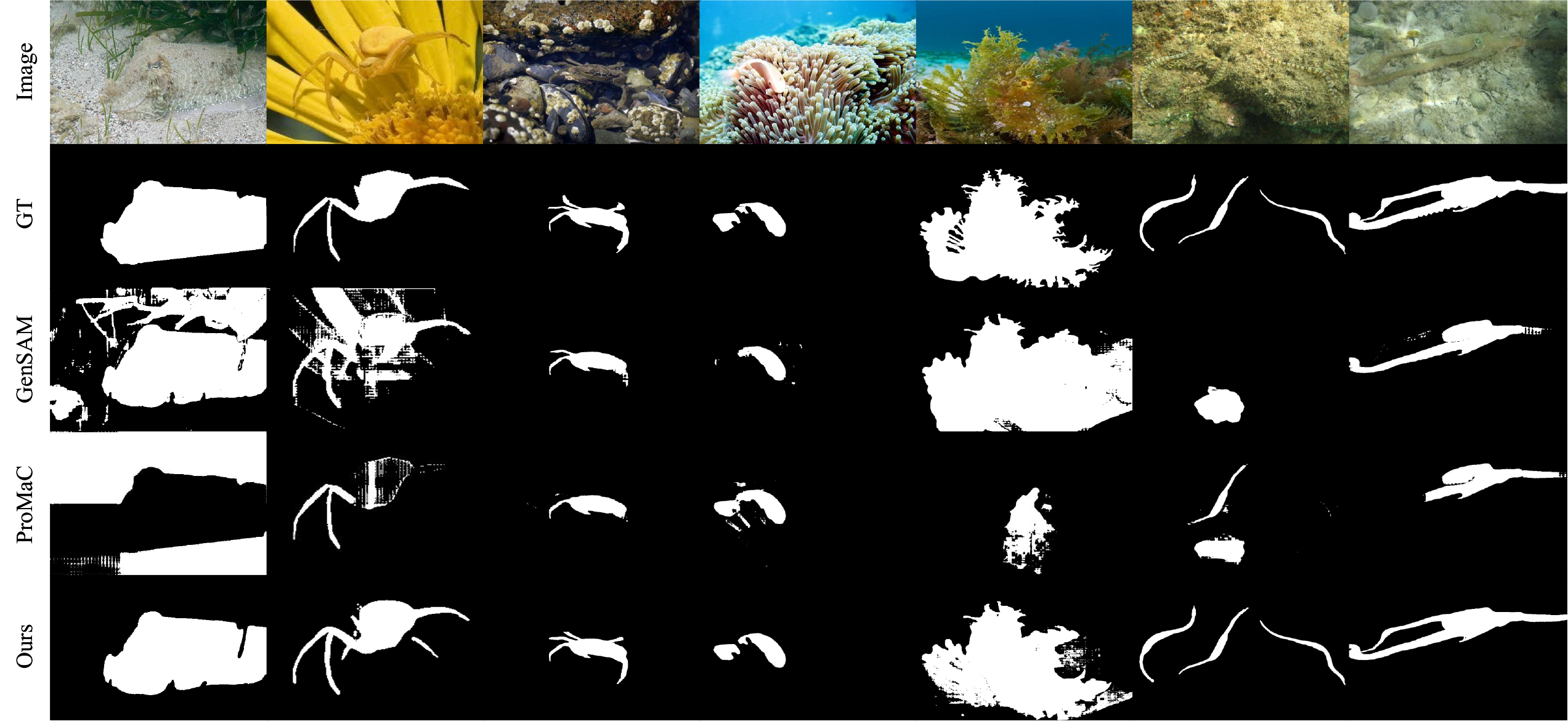}
\caption{Qualitative results on the COD10K dataset.}
\label{figure8}
\end{figure}

\begin{figure}[tb]
\centering
\includegraphics[width=1\textwidth]{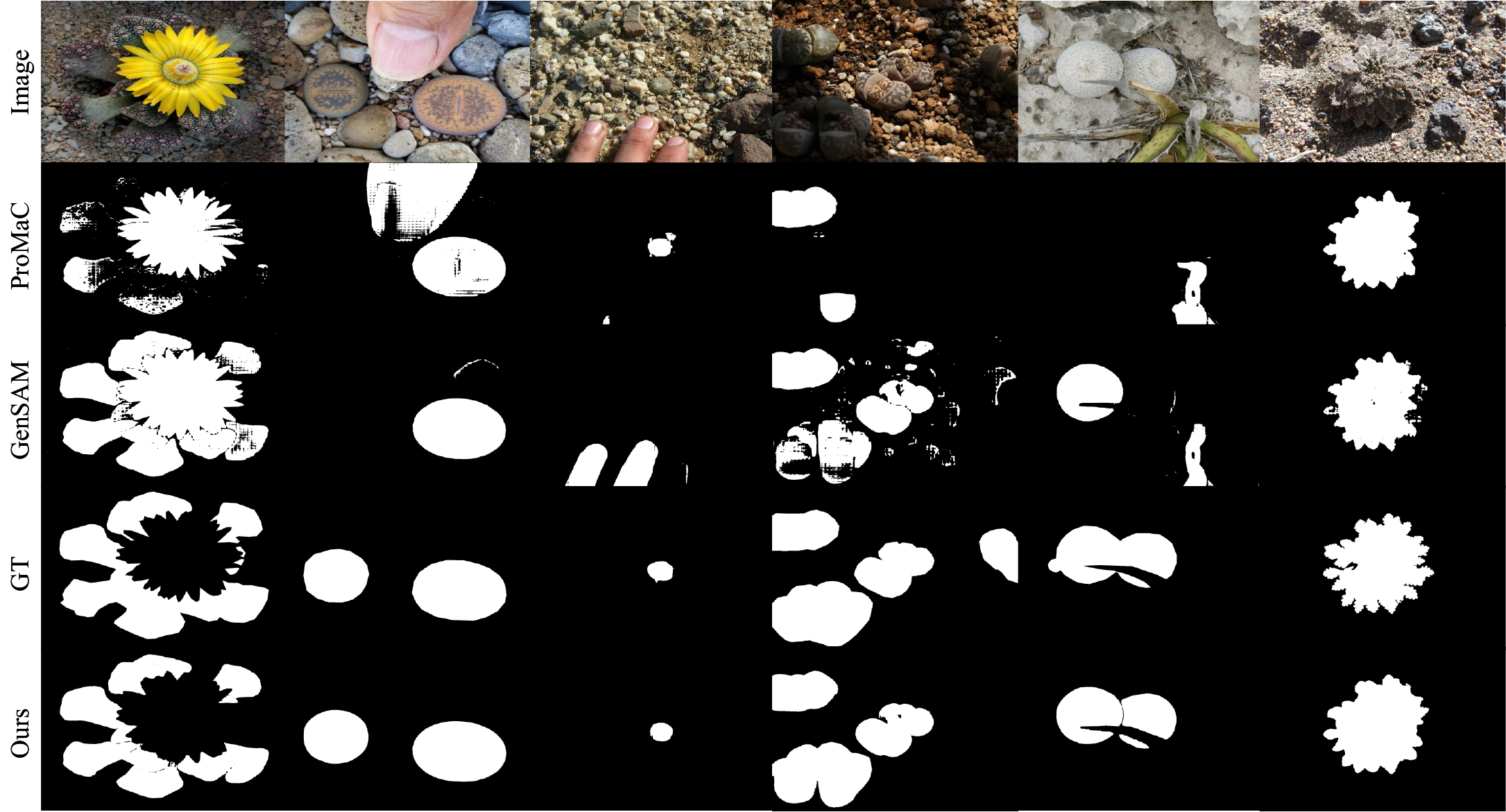}
\caption{Qualitative results on the PlantCAMO dataset.}
\label{figure9}
\end{figure}

\begin{figure}[tb]
\centering
\includegraphics[width=1\textwidth]{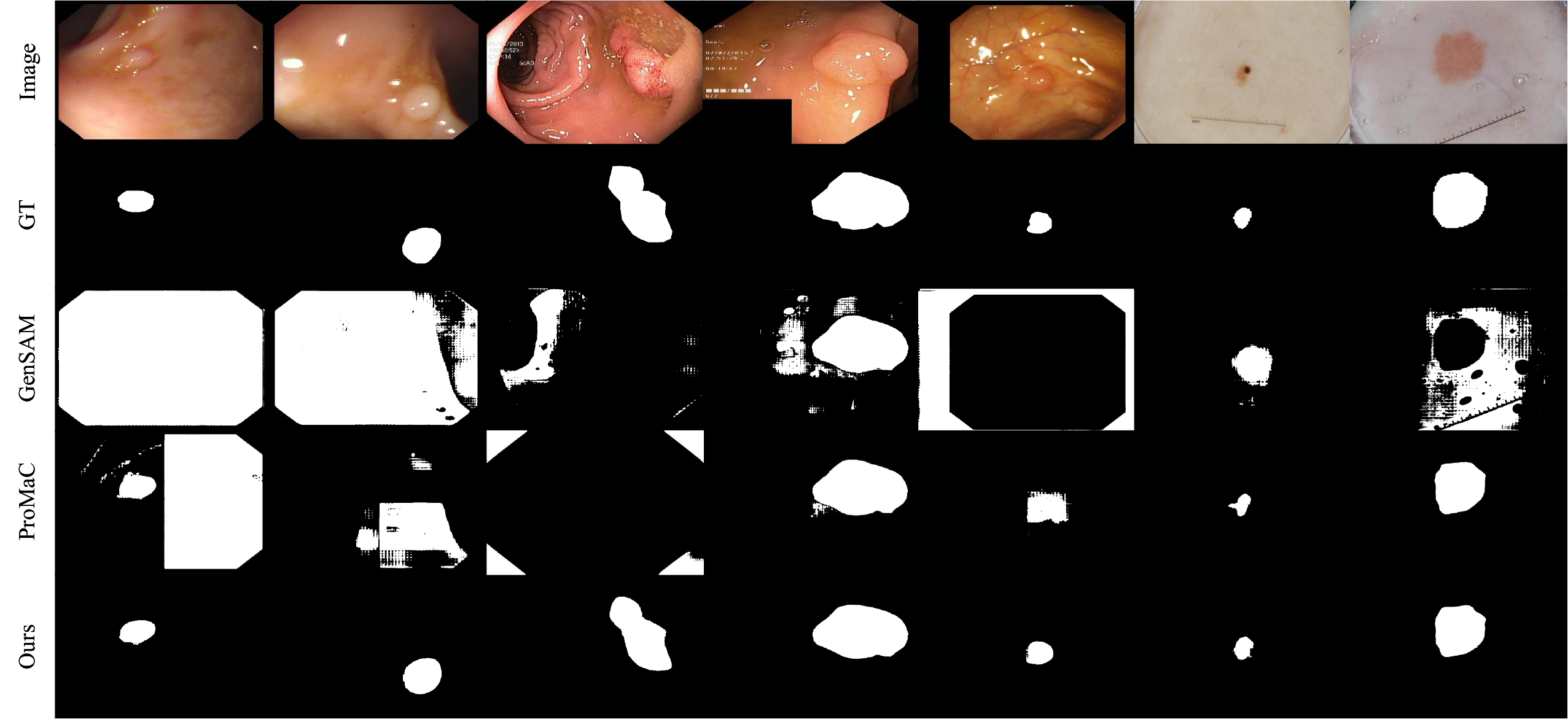}
\caption{Qualitative results on the MIS task dataset.}
\label{figure10}
\end{figure}

Figures~\ref{figure6}--\ref{figure10} illustrate the qualitative superiority of \textit{ArgusCogito} under fully unsupervised, zero-shot conditions. Compared to existing methods like GenSAM and ProMAC, which often struggle with structural misalignment or incomplete segmentation, our framework consistently delivers coherent and precise predictions, preserving object boundaries and semantic integrity.

On the CAMO dataset (Fig.~\ref{figure6}), which features highly camouflaged objects in visually homogeneous scenes, both GenSAM and ProMAC tend to produce fragmented or noisy masks. GenSAM, relying on weak semantic heatmaps generated from generic visual-text alignments, frequently misfires in regions of low saliency, while ProMAC’s reliance on independently cropped patches limits its ability to perceive spatial continuity—often leading to under-segmentation. In contrast, \textit{ArgusCogito} achieves superior boundary precision and shape completeness, benefiting from its global, full-resolution reasoning and iterative refinement stages that progressively improve mask fidelity.

In the CHAMELEON dataset (Fig.~\ref{figure7}), where foreground-background entanglement is particularly severe due to subtle texture variations and minimal luminance differences, segmentation becomes especially prone to false positives. GenSAM often hallucinates irrelevant regions, and ProMAC fails to capture global structure, producing disconnected or missing parts. \textit{ArgusCogito}, however, maintains structural coherence, leveraging its dynamic spatial focus and sculpting mechanisms to filter out distractors while attending to semantically relevant, yet visually elusive, regions.

The COD10K dataset (Fig.~\ref{figure8}) offers a broad spectrum of object classes and natural backgrounds, presenting challenges in terms of semantic generalization and category mismatch. Existing prompt-based approaches often suffer from semantic drift or over-reliance on specific training distributions, which leads to performance drops in unseen categories. \textit{ArgusCogito} exhibits strong robustness here: guided by a task-generic prompt and a structured reasoning chain, it maintains high-quality segmentation even for rare or structurally ambiguous classes.

On the PlantCAMO dataset (Fig.~\ref{figure9}), which introduces a domain shift with fine-grained plant camouflage patterns and inter-class similarity, the shortcomings of generic visual-language models become more pronounced. GenSAM and ProMAC, lacking exposure to this specialized context, fail to produce discriminative masks, often merging foreground and background due to insufficient visual priors. Remarkably, \textit{ArgusCogito} achieves consistent delineation without any domain-specific adaptation, demonstrating its strong cross-domain generalization ability and sensitivity to fine structural cues—thanks to its multi-stage design that aligns geometry, semantics, and spatial layout.

We further present qualitative results on two representative medical image segmentation (MIS) tasks under zero-shot settings: polyp segmentation (CVC-ColonDB and Kvasir) and skin lesion segmentation (ISIC), as illustrated in Fig.~\ref{figure10}. These domains are particularly challenging due to low-contrast object boundaries, flattened image geometry, high inter-class similarity, and the lack of consistent shape or texture priors.

Compared with camouflaged natural scenes, the object of interest in medical images often occupies a near-circular or amorphous region within a visually homogeneous background (e.g., black endoscopic surroundings or uniform dermoscopic fields). This structural simplicity paradoxically increases segmentation difficulty for prompt-based methods that rely heavily on generic saliency or context priors. GenSAM, for instance, frequently misclassifies the dark peripheral background as foreground, while entirely neglecting the actual lesion or polyp. This failure stems from its weak semantic grounding and over-reliance on low-level saliency cues derived from generic visual-language alignment. Similarly, ProMAC's patch-wise masking mechanism—designed for natural scenes with diverse textures—tends to disregard global object continuity. As a result, it either fragments the target or suppresses it altogether, especially when the object does not conform to expected saliency or occupies uniform intensity regions.

In contrast, \textit{ArgusCogito} demonstrates remarkable robustness despite no exposure to medical imagery during training. Its prompt-informed spatial reasoning, combined with iterative sculpting, enables precise delineation even in cases with low texture contrast and ambiguous boundaries. The framework effectively discerns subtle structural cues and preserves spatial coherence, highlighting its strong cross-domain generalization and semantic adaptability.

These visualizations further substantiate the quantitative results and validate the framework’s capacity to segment complex, low-saliency targets in unfamiliar domains—ranging from natural camouflage to clinical images—without any task-specific adaptation.
\section*{E Discussion}
\label{sec:E}
ArgusCogito demonstrates strong generalization across diverse segmentation tasks without task-specific training, indicating significant potential for broader real-world deployment. To enhance its robustness and versatility, key technical directions merit exploration:

\begin{itemize}
    \item \textbf{Multimodal Integration}  
    Incorporate complementary modalities (e.g., infrared/thermal imaging) to mitigate limitations of RGB/monocular depth cues in low-light or occluded environments. Multimodal fusion boosts perceptual redundancy and tolerance to sensor degradation, enabling reliable operation in safety-critical domains (e.g., search-and-rescue, night-time surveillance).  

    \item \textbf{Adaptive Reasoning with Granularity Awareness}  
    Implement dynamic inference adjustments: local zoom-in for ultra-small target localization (based on estimated object size/density) and hierarchical decomposition + structured fusion for dense-object regions to preserve global consistency.  

    \item \textbf{Domain Adaptability Enhancements}  
    Leverage lightweight fine-tuning (minimal supervision) to inject task-specific knowledge into the vision-language backbone. Alternatively, introduce a dynamic camouflage-pattern lexicon to expand semantic priors, aiding interpretation of complex visual semantics in specialized domains (e.g., underwater ecosystems, military concealment).  

    \item \textbf{Scenario Expansion}  
    Extend the framework to low-contrast segmentation scenarios (e.g., weak-feature defect detection in industrial inspection) to validate generality. Further, advance to temporally consistent video segmentation, capturing inter-frame motion trajectories to model dynamic camouflage behaviors.  
\end{itemize}

These directions outline a pathway toward a more generalizable, modality-resilient, and context-aware visual segmentation system — laying a foundation for future progress in open-world perception and task-adaptive visual reasoning.

\end{document}